\documentclass[11pt]{elsarticle}

\usepackage{lineno,hyperref}

\usepackage{float}
\usepackage{graphicx}
\usepackage{subfig}
\usepackage{mathrsfs}
\usepackage{graphics}
\usepackage{float}
\usepackage{amsmath}
\usepackage{amsfonts}
\usepackage{mathrsfs}
\usepackage{amsmath}

\usepackage{optidef}

\usepackage{optidef}
\usepackage{graphicx}
\usepackage{float}
\usepackage{graphicx}
\usepackage{subfig}
\usepackage{mathrsfs}
\usepackage{float}
\usepackage{amsmath}
\usepackage{amsfonts}
\usepackage{booktabs}
\usepackage{natbib}
\usepackage{bbm}
\usepackage{algorithm}
\usepackage[noend]{algpseudocode}
\usepackage{multirow}
\algdef{SE}[DOWHILE]{Do}{doWhile}{\algorithmicdo}[1]{\algorithmicwhile\ #1}

\usepackage[letterpaper, portrait, margin=1in]{geometry}

\usepackage{booktabs}
\usepackage{tabularx}

\usepackage{xcolor}
\hypersetup{
    colorlinks,
    linkcolor={red!50!red},
    citecolor={blue!50!blue},
    urlcolor={blue!80!blue}
}

\usepackage{optidef}

\usepackage{amsthm}

\theoremstyle{definition}

\usepackage{newtxtext} 

\newcommand{\sfunction}[1]{\textsf{\textsc{#1}}}
\algrenewcommand\algorithmicforall{\textbf{foreach}}
\algrenewcommand\algorithmicindent{.8em}

\DeclareMathOperator*{\argmax}{arg\,max}
\DeclareMathOperator*{\argmin}{arg\,min}

\makeatletter
\newcommand*{\centerfloat}{%
  \parindent \z@
  \leftskip \z@ \@plus 1fil \@minus \marginparwidth
  \rightskip \leftskip
  \parfillskip \z@skip}
\makeatother

\usepackage{amssymb,amsmath,caption}
\usepackage[hang,flushmargin]{footmisc}
\newcommand{\algorithmfootnote}[2][\footnotesize]{
  \let\old@algocf@finish\@algocf@finish
  \def\@algocf@finish{\old@algocf@finish
    \leavevmode\rlap{\begin{minipage}{\linewidth}
    #1#2
    \end{minipage}}%
  }%
}

\usepackage{setspace}
\usepackage{natbib}
\usepackage{algorithm}
\usepackage[noend]{algpseudocode}
\algdef{SE}[DOWHILE]{Do}{doWhile}{\algorithmicdo}[1]{\algorithmicwhile\ #1}

\usepackage{appendix}

\usepackage{array}
\usepackage{threeparttable}
\usepackage{multirow}
\usepackage{rotating}
\usepackage{makecell}
\usepackage{longtable}
\usepackage[nolist]{acronym}

\journal{}

\biboptions{authoryear}

\begin{document}
\begin{frontmatter}

\title{Predicting Drivers' Route Trajectories in Last-Mile Delivery Using A Pair-wise Attention-based Pointer Neural Network}


\author[label1]{Baichuan Mo}
\author[label1]{Qing Yi Wang\corref{mycorrespondingauthor}}
\author[label1]{Xiaotong Guo}
\author[label3]{Matthias Winkenbach}
\author[label4]{Jinhua Zhao}
\address[label1]{Department of Civil and Environmental Engineering, Massachusetts Institute of Technology, Cambridge, MA 02139}
\address[label3]{Center for Transportation and Logistics, Massachusetts Institute of Technology, Cambridge, MA 20139}
\address[label4]{Department of Urban Studies and Planning, Massachusetts Institute of Technology, Cambridge, MA 20139}


\begin{abstract}
In last-mile delivery, drivers frequently deviate from planned delivery routes because of their tacit knowledge of the road and curbside infrastructure, customer availability, and other characteristics of the respective service areas. Hence, the actual stop sequences chosen by an experienced human driver may be potentially preferable to the theoretical shortest-distance routing under real-life operational conditions. Thus, being able to predict the actual stop sequence that a human driver would follow can help to improve route planning in last-mile delivery. This paper proposes a pair-wise attention-based pointer neural network for this prediction task using drivers' historical delivery trajectory data. In addition to the commonly used encoder-decoder architecture for sequence-to-sequence prediction, we propose a new attention mechanism based on an alternative specific neural network to capture the local pair-wise information for each pair of stops. To further capture the global efficiency of the route, we propose a new iterative sequence generation algorithm that is used after model training to identify the first stop of a route that yields the lowest operational cost. Results from an extensive case study on real operational data from Amazon's last-mile delivery operations in the US show that our proposed method can significantly outperform traditional optimization-based approaches and other machine learning methods (such as the  Long Short-Term Memory encoder-decoder and the original pointer network) in finding stop sequences that are closer to high-quality routes executed by experienced drivers in the field. Compared to benchmark models, the proposed model can increase the average prediction accuracy of the first four stops from around 0.229 to 0.312, and reduce the disparity between the predicted route and the actual route by around 15\%.
\end{abstract}

\begin{keyword}
Route planning, Trajectory prediction, Sequence-to-sequence model, Last-mile delivery, Pointer network, Attention
\end{keyword}

\end{frontmatter}


\begin{acronym}
\acro{TSP}{traveling salesman problem}
\acro{VRP}{vehicle routing problem}
\acro{ASNN}{alternative specific neural network}
\acro{LSTM}{Long Short-Term Memory}
\acro{LSTM-E-D}{LSTM-encoder-decoder}
\acro{seq2seq}{sequence-to-sequence}
\acro{RNN}{recurrent neural network}
\acro{CNN}{convolutional neural network}
\acro{RL}{reinforcement learning}
\acro{SL}{supervised learning}
\end{acronym}

\section{Introduction}
\label{sec:intro}

The optimal planning and efficient execution of last-mile delivery routes is becoming increasingly important for the business operations of many logistics service providers around the globe for a variety of reasons.
E-commerce volumes are growing rapidly and make up a constantly growing share of overall retail sales. For instance, in the US, the share of e-commerce sales in total retail sales has grown from around 4\% in 2010 to around 13\% in 2021. Even by the end of 2019, i.e., before the outbreak of the COVID-19 pandemic, it had reached 11\% \citep{USCensusBureau2021Quarterly2021}. 
Undoubtedly, the pandemic further accelerated the growth of e-commerce \citep{postnord2021E-commerceEurope,McKinseyCompany2021HowCountries}.
In the medium to long run, its growth will continue to be fueled by an ongoing trend towards further urbanization, which is particularly pronounced in developing and emerging economies \citep{UnitedNationsDepartmentofEconomicandSocialAffairs2019WorldST/ESA/SER.A/420}.
The share of the global population living in urban areas is currently projected to rise from around 55\% in 2018 to around 68\% by 2050.
The associated increase in population density in most urban areas will likely lead to growing operational uncertainties for logistics service providers, as increasing congestion levels, less predictable travel times, and scarce curb space make efficient and reliable transport of goods into and out of urban markets increasingly challenging \citep{Rose2016ExploringProviders}.

As a result of the continued boom of e-commerce and constantly growing cities, global parcel delivery volumes have been increasing rapidly in recent years and are expected to continue to do so. Across the 13 largest global markets, including the US, Brazil, and China, the volume of parcels delivered more than tripled from 43 billion in 2014 to 131 billion in 2020 \citep{PitneyBowes2020PitneyIndex}. 
At the same time, customer expectations towards last-mile logistics services are rising. For instance, there is a growing demand for shorter delivery lead times, including instant delivery services and same-day delivery, as well as customer-defined delivery preferences when it comes to the time and place of delivery \citep{lim2019configuring,Cortes2021Last-mileIndustry,snoeck2021discrete}. 

When applied to realistically sized instances of a last-mile delivery problem, solving the underlying \ac{TSP} or \ac{VRP} to (near) optimality becomes challenging, as both problem classes are known to be NP-hard.
Traditional \ac{TSP} and \ac{VRP} formulations aim to minimize the total distance or duration of the route(s) required to serve a given set of delivery stops. The operations research literature has covered the \ac{TSP}, \ac{VRP}, and their many variants extensively, and in recent years important advances have been made with regards to solution quality and computational cost. However, in practice, many drivers, with their own tacit knowledge of delivery routes and service areas, divert from seemingly optimal routes for reasons that are difficult to encode in an optimization model directly. For example, experienced drivers may have a better understanding of which roads are hard to navigate, at which times traffic is likely to be bad, when and where they can easily find parking, and which stops can be conveniently served together. Therefore, compared to the theoretically optimal (i.e., distance or time minimizing) route, the \emph{deviated actual route sequence} chosen by an experienced human driver is potentially preferable under real-life operational conditions. 
Due to their often simplistic, single-objective nature, most theoretical route optimization algorithms fail to capture the complex, experience-based decision-making of delivery drivers, who would judge the quality of a route sequence not just based on a single metric such as time or distance, but by implicitly factoring in criteria such as convenience, efficiency, and safety \citep{merchan20222021}.
This notion has been confirmed by anecdotal evidence of industry experts from a variety of leading last-mile carriers. While these players continue to invest heavily into developing new and better route optimization algorithms, they all allow their experienced drivers to deviate from the theoretical route plans if necessary, in order to uncover further potential for algorithm improvement and learning from operational practice \citep[see, e.g.,][]{fastcompany2013UPS}.

An important challenge in today's last-mile delivery route planning is therefore to leverage historical route execution data to propose planned route sequences that are close to the actual trajectories that would be executed by drivers, given the delivery requests and their characteristics. Note that, while distance and time-based route efficiency is still an important factor for planning route sequences, it is not the sole objective, as tacit driver knowledge is also incorporated in the proposed route sequences. 
Unlike a typical \ac{VRP} in which the number of vehicles and their respective route sequences need to be determined simultaneously, in this study, we focus on solving a problem that is similar to a \ac{TSP} at the individual vehicle level. That is, we aim to solve a stop sequence to serve a given set of delivery requests, and expect that the proposed stop sequence is as close to the actual trajectories that would be executed by drivers as possible. 

To this end, we propose a pair-wise attention-based pointer neural network to predict the actual route sequence taken by delivery drivers using drivers' historical delivery trajectory data. The proposed model follows a typical encoder-decoder architecture for the sequence-to-sequence prediction. However, unlike previous studies, we propose a new attention mechanism based on an \ac{ASNN} to capture the local pair-wise information for each stop pair. To further capture the global efficiency of the route (i.e., its operational cost in terms of total distance or duration), after model training, we propose a new sequence generation algorithm that iterates over different first stops and selects the route with the lowest operational cost. 

The main contribution of this paper is three-fold:
First, we propose a new \ac{ASNN}-based attention mechanism to capture the local information between pairs of stops (e.g., travel time, geographical relation), which can be well adapted to the original pointer network framework for sequence prediction. Second, we propose a new sequence generation algorithm that iterates over different first stops in the predicted route sequences and selects the lowest operational cost route. The intuition is that the stop-to-stop relationship (referred to as the \emph{local view}) is easier to learn from data than the stop sequence of the route as a whole (referred to as the \emph{global view}). 
Lastly, we apply our proposed method to a large set of routes executed by Amazon delivery drivers in the US. The results show that our proposed model can outperform traditional optimization-based approaches and other machine learning methods in finding stop sequences that are closer to high-quality routes executed by experienced drivers in the field. 

The remainder of this paper is structured as follows. 
In Section \ref{sec:prob_set} we define the problem setting under investigation in a more formal way.
Section \ref{sec_liter} then reviews previous studies in the literature related to this paper. 
Section \ref{sec_method} presents our methodology and elaborates on the detailed architecture of the proposed pair-wise attention-based pointer neural network. 
Section \ref{sec_case} presents the experimental setup and numerical results of our case study, applying our proposed method to real-world data made available by the Amazon Last-Mile Routing Research Challenge \citep{merchan20222021, Winkenbach2021TechnicalChallenge}. 
Section \ref{sec_dis_con} concludes this paper and discusses future research directions.

\section{Problem Setting}
\label{sec:prob_set}

In the last-mile delivery routing problem considered here, a set of stops $\mathcal{S} = \{s_1,...,s_n\}$ to be served by a given delivery vehicle is given to the route planner. 
The planner's objective is to find the optimal stop sequence that has the minimal operational cost. In this case, we consider total cost as total travel time.
The planner is given the expected operational cost (i.e., travel times) between all pairs of stops $(s_i,s_j)$.
The theoretically optimal stop sequence, denoted by $(s_{(1)}^{\text{T}}, ..., s_{(n)}^{\text{T}})$, can be found by solving a \ac{TSP} formulation. This stop sequence is referred to as the planned stop sequence.
However, as discussed in Section \ref{sec:intro}, minimizing the theoretical operational cost (i.e., total travel time) of the route may not capture experienced drivers' tacit knowledge about the road network, infrastructure, and recipients. 
Therefore, the actual driver executed stop sequence $(s_{(1)}, ..., s_{(n)})$ can be different from the planned route sequence. Note that here, $s_{(i)} \in \mathcal{S}$ denotes the $i$-th stop that is actually visited by the driver.

The objective of the model presented in this study is to predict the actual driver executed sequence $(s_{(1)}, ..., s_{(n)})$ given a set of stops $\mathcal{S}$ and the corresponding delivery requests and characteristics $X^{\mathcal{S}}$ (such as the number of packages, estimated service time for each package, geographical information for each stop, travel time between each stop pairs, etc.).
All drivers are assumed to start their routes from a known depot $D^{\mathcal{S}}$ and return back to $D^{\mathcal{S}}$. Therefore, the complete trajectory should be a tour $(D^{\mathcal{S}}, s_{(1)}, ..., s_{(n)}, D^{\mathcal{S}})$. For the convenience of model description, we ignore the depot station in the sequence.

\begin{figure}[htb]
    \centering
    \includegraphics[width = 0.8 \linewidth]{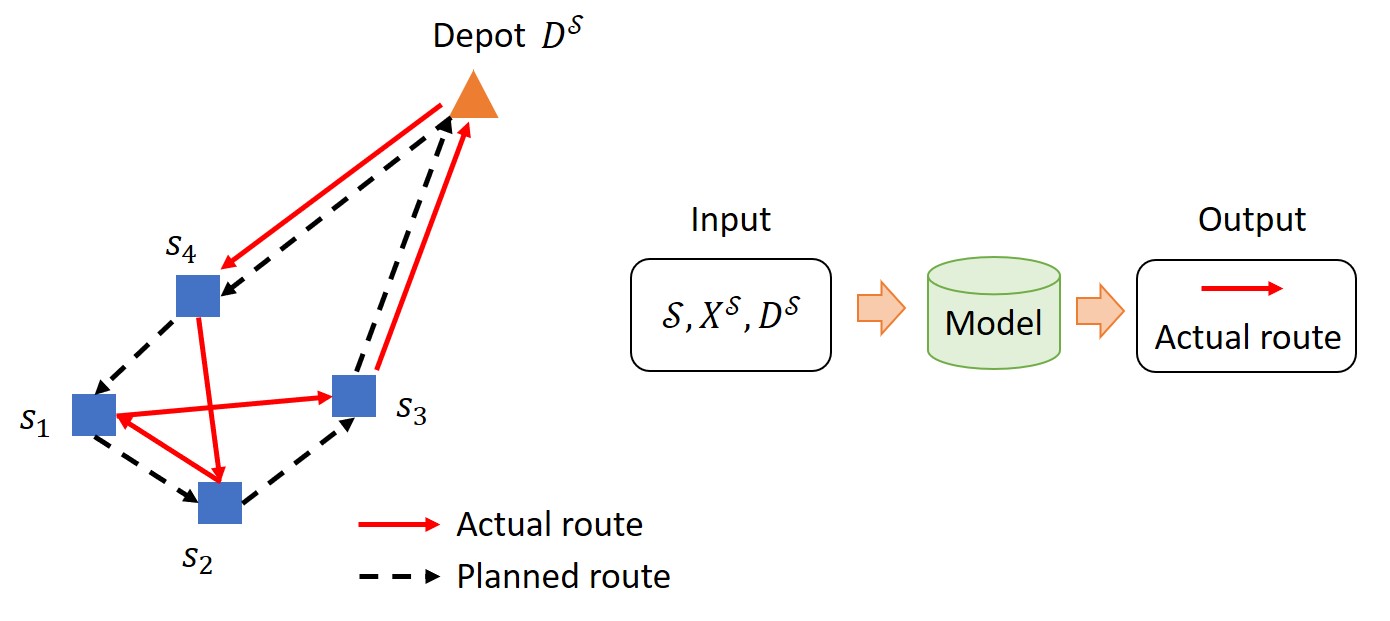}
    \caption{Illustrative example of the problem setting.}
    \label{fig_example}
\end{figure}

Figure \ref{fig_example} provides a simple example for illustration. In this example, we are given four stops $\mathcal{S} = \{s_1,s_2,s_3,s_4\}$ and a depot $D^{\mathcal{S}}$. The planned stop sequence for the driver is $(s_4,s_1,s_2,s_3)$, while the actual stop sequence executed by the driver is $(s_4,s_2,s_1,s_3)$. The proposed model aims to predict the actual sequence $(s_4,s_2,s_1,s_3)$ given the depot location $D^{\mathcal{S}}$, the set of stops to be visited $\mathcal{S}$, and characteristics of the stops $X^{\mathcal{S}}$.
This problem setup is inspired by the Amazon Last-Mile Routing Research Challenge \citep[cf., ][]{Winkenbach2021TechnicalChallenge}. Note that this study only focuses on the stop sequence prediction. The routing between stops is not considered. It is assumed that the drivers always take the optimal route between stops, which is reflected by the travel time matrix between stops in our problem setup.

\section{Literature Review}\label{sec_liter}

The problem setting defined in Section \ref{sec:prob_set} involves both solving a cost-minimizing routing problem (i.e., the \ac{TSP}) and capturing tacit driver knowledge to learn systematic deviation of drivers from the planned and theoretically optimal stop sequences.  Therefore, we will first review the extant literature on the \acp{TSP} and its most relevant variants. We will then go through various machine learning approaches that have been proposed by the extant literature to generate sequences, with a section on methods specifically for solving the \ac{TSP}. Note that although these machine learning approaches are used to solve the \ac{TSP} instead of the actual routes taken by drivers, their architectures may be helpful to learn the actual route as well.   

\subsection{Travelling salesman problems}

First, given the travel times between stops, a solution to the \ac{TSP}, which finds the route with the minimum cost or distance (i.e., the planned route), can be a close approximation of the actual route. Since the drivers are paid for the number of packages delivered, all drivers' goal is to deliver the packages in the minimum amount of time. Most of the drivers do follow large parts of the planned routes.

The \ac{TSP} is a well-known NP-hard problem that has been studied extensively over the last century, with a lot of books and review papers published on its history, formulations, solution approaches, and applications \citep{Applegate2006TheProblem, Matai2010, Davendra2020}. An overview of the relevant \ac{TSP} variants and solution approaches are presented below.

The basic setup of \ac{TSP} has one traveler and requires the traveler to return to the starting point after visiting each node exactly once, and that the traveling cost matrix (represented by distance and/or time) is symmetric (cost between $i$ and $j$ is the same with that between $j$ and $i$). In most real-world applications, the basic setup needs to be modified. For example, the cost matrix, if represented by travel times, is likely asymmetric. This variant of \ac{TSP} is thus named asymmetric \ac{TSP} (ATSP) \citep{Jonker1983}. In some applications, the vehicle does not need to return to the original depot \citep{Traub2021}, or it can charge/refuel and potentially load additional delivery items at intermediate stops \citep{Kkolu2019}. In many last-mile delivery applications, some packages are time-sensitive, and therefore time window constraints to their delivery need to be considered in a so-called \ac{TSP} with time windows (TSPTW) \citep{daSilva2010, Mladenovi2012}. In large systems, there might be more than one salesman serving a set of stops, resulting in multiple traveling salesmen problems (MTSPs) \citep{Cheikhrouhou2021}. 

Different variants of \ac{TSP} further impose different constraints on the solution. While some problems can be reduced to the basic setup in the formulation stage, others require more versatile solution algorithms. In general, the solution approaches to the \ac{TSP} can be divided into exact approaches and approximate approaches. Exact approaches include branch-and-cut \citep{yuan2020branch} and branch-and-bound \citep{salman2020branch}. Since the \ac{TSP} is a well-known NP-hard problem, exact approaches can only be applied on problems of smaller scale, or aid in heuristics to cut the solution space. Among approximate approaches, there are heuristics designed for the \ac{TSP} specifically, as well as meta-heuristics that are generic and treat the problem like a blackbox. The most commonly used heuristics and meta-heuristics include nearest neighbor searches, local searches, simulated annealing, and genetic algorithms. A more comprehensive review of existing solution approaches can be found in \citet{Halim2017, Purkayastha2020}.
Despite the \ac{TSP} being NP-hard, modern mixed-integer optimization solvers (e.g., Gurobi, CPLEX, or GLPK) can solve it efficiently for real-world instances  by combining exact approaches with heuristics~\citep{GUO2022103691}.


\subsection{\Acl{seq2seq} prediction using deep learning}

The \ac{TSP} and its variants are a viable option for sequence generation only when the objective is clearly defined. They fall short when the sequence generation problem does not have a well-defined cost-minimization objective. In a lot of applications, the rule of sequence generation cannot be simply defined and optimized.

A standard example for a sequence learning problem is machine translation, where a sequence of words in one language needs to be translated to another language. Another type of sequence learning is time series modeling, where a sequence of historical observations is given to predict future states of the system. In both cases, the primary modeling task is to learn the sequence generation rules. In recent years, deep learning has successfully achieved great performance in various settings of sequence generation. These models are often referred to as \acf{seq2seq} models.

\Ac{seq2seq} models often consist of an encoder and a decoder, where the encoder encodes the input sequence into a fixed-length vector representation, and the decoder generates a sequence based on the generated vector representation. 
Most encoder-decoder architectures adopt \ac{RNN} layers and its variants such as \ac{LSTM} \citep{hochreiter1997long} and gated recurrent layers (GRU) \citep{cho2014learn} to learn long-range dependencies. Early works using \ac{LSTM} alone were able to generate plausible texts  \citep{Graves2013GeneratingNetworks} and translate between English and French \citep{Sutskever2014SequenceNetworks} with long-range dependencies. \citet{chung2014empirical} demonstrate the superiority of GRU compared to \acp{LSTM} in music and speech signal modeling.

Attention-based mechanisms, first introduced by \citet{Bahdanau2015NeuralTranslate}, have been shown to be a great addition since it allows the decoder to selectively attend to parts of the input sequence and relieves the encoder of the task of encoding all the information into a fixed-length vector representation. Most sequence generation problems benefit from keeping track of long-range dependencies and global context while decoding. To address that, multi-level attention was proposed to capture the local and global dependency, and has shown to be effective in speech recognition \citep{chorowski2015attention}, text generation \citep{liu2018table}, and machine translation tasks \citep{luong2015effective}.

The encoder-decoder architecture combined with attention is very versatile, and it can be combined with other deep learning architectures to perform sequence learning in addition to language tasks. The \ac{LSTM} and attention architecture is applied to semantic trajectory prediction \citep{karatzoglou2018aseq2seq}, text summarization \citep{liang2020abstractive}, demand modelling \citep{REN2020101834}, and wind power forecasting \citep{zhang2020short-term}. When the goal is set to recover the original sequence, unsupervised learning of molecule embedding can be obtained for downstream classification tasks \citep{xu2017seq2seq}. When the spatial dimension is added, a \ac{CNN} layer can be added, and the dimension of the sequence generated can be expanded. For example, \citet{wang2020seqst-gan} predict a city's crowd flow patterns, and \citet{wu2020pq-net} generate 3D shapes via sequentially assembling different parts of the shapes.




While \ac{RNN}-based architectures are still a widely adopted choice for \ac{seq2seq} modeling, attention can also be used as a standalone mechanism for \ac{seq2seq} translations independent of \acp{RNN}. The idea was proposed by \citet{Vaswani2017} in an architecture named transformer. Without recurrence, the network allows for significantly more parallelization, and is shown to achieve superior performance in experiments, and powered the popularity of transformer-based architectures in various sequence generation tasks \citep{huang2018music,lu2021context}. A separate line of work by \citet{zhang2019abstract} also demonstrated that a hierarchical \ac{CNN} model with attention outperforms the traditional \ac{RNN}-based models.

\subsection{Using deep learning to generate \ac{TSP} solutions}

The above \ac{seq2seq} translation mechanisms work well when the input data is naturally organized as a sequence, and the output sequence corresponds to the input sequence, such as in music and language. However, in our paper, the input is an unordered sequence, and the output has the same but re-ordered elements of the same input sequence. In this case, the concept of attention is helpful and has been successfully used to produce solutions to the \ac{TSP}. The pointer network, proposed by \cite{Oriol2015} and further developed in \cite{Vinyals2016}, uses attention to select a member of the input sequence at each decoder step. While it is not required that the input sequence is ordered, an informative ordering could improve the performance \citep{Vinyals2016}.

While the original pointer network was solved as a classification problem and cross-entropy loss was used, it is not necessarily the most efficient choice. The cross-entropy loss only distinguishes between a correct prediction and an incorrect prediction. But in instances like routing, the distances between the predicted position and the correct position, as well as the ordering of subsequences, could incur different costs in practice. Further developments in solving \ac{TSP} with machine learning methods involve \ac{RL}, which enables the optimization of custom evaluation metrics \citep{Bello2019, Kool2019, Ma2019,LIU2020102070}. \citet{Joshi2019} compared the performance of \ac{RL} and \ac{SL} on \ac{TSP} solutions and found that \ac{SL} and \ac{RL} models achieve similar performance when the graphs are of similar sizes in training and testing, whereas \ac{RL} models have better generalizability over variable graph sizes. However, \ac{RL} models require significantly more data points and computational resources, which is not always feasible.



Although this \ac{seq2seq} and attention framework has only been used to reproduce \ac{TSP} solutions, it provides an opportunity to learn and incorporate additional information beyond the given travel times and potentially learn individual differences when more information is given to the neural network. 
In this paper, we combine the ideas of \ac{seq2seq} modeling and attention to predict the actual route executed by a driver. 


\section{Methodology}\label{sec_method}

This section details the methodology proposed to address the problem. First, the high-level \ac{seq2seq} modelling framework is introduced, followed by the explanation of the novel pair-wise attention and sequence generation and selection mechanism used within the modelling framework.

\subsection{\Acl{seq2seq} modeling framework}
Let the input sequence be an arbitrarily-ordered sequence $(s_1,...,s_n)$. Denote the output sequence as $(\hat{s}_{(1)}, ..., \hat{s}_{(n)})$. Let $c_i$ indicate the ``position index'' of stop $\hat{s}_{(i)}$ with respect to the input sequence (where $c_i \in \{1,...,n\}$). For example, for input sequence $(B,A,C)$ and output sequence $(A, B, C)$, we have $c_1 = 2$, $c_2 = 1$, $c_3 = 3$, which means the first output stop $A$ is in the second position of the input sequence $(B,A,C)$ and so on. 

The \ac{seq2seq} model computes the conditional probability $\mathbb{P}(c_1,...,c_n \mid {\mathcal{S}}; \theta)$ using a parametric neural network (e.g., recurrent neural network) with parameter $\theta$, i.e.,
\begin{align}
    \mathbb{P}(c_1,...,c_n \mid {\mathcal{S}}, X^{\mathcal{S}}; \theta) = \mathbb{P}(c_1 \mid {\mathcal{S}}, X^{\mathcal{S}}; \theta) \cdot \prod_{i = 2}^{n}\mathbb{P}(c_i \mid c_1,...,c_{i-1}, {\mathcal{S}}, X^{\mathcal{S}}; \theta)
\end{align}

The parameters of the model are learnt by empirical risk minimization (maximizing the conditional probabilities on the training set), i.e.,
\begin{align}
    \theta^* = \argmax_{\theta} \sum_{\mathcal{S}} \mathbb{P}(c_1,...,c_n \mid {\mathcal{S}}, X^{\mathcal{S}}; \theta)
    \label{eq_obj_LL}
\end{align}
where the summation of $\mathcal{S}$ is over all training routes. In the following section, we will elaborate how $\mathbb{P}(c_i \mid c_1,...,c_{i-1}, {\mathcal{S}}, X^{\mathcal{S}}; \theta)$ is calculated using the pair-wise attention-based pointer neural network.

\subsection{Pair-wise attention-based pointer neural network}
\begin{figure}[htb]
\centering
{\includegraphics*[width = 0.8 \linewidth]{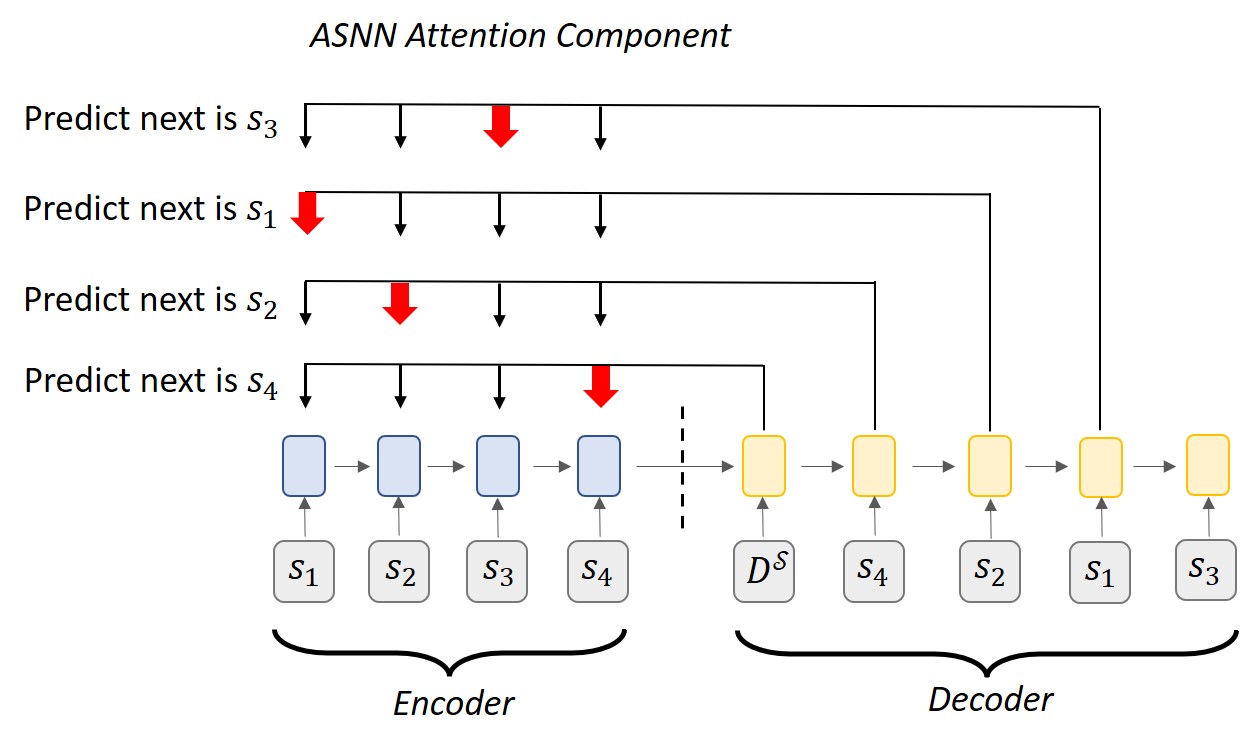}}
\caption{Overall architecture of the pair-wise attention-based pointer neural network (adapted from \citet{Oriol2015}) \label{fig_arch_example}}
\end{figure}

Figure \ref{fig_arch_example} uses a four-stop example to illustrate the architecture of the proposed model. The whole model is based on the \ac{LSTM} encoder and decoder structure. In particular, we use one \ac{LSTM} (i.e., encoder) to read the input sequence, one time step at a time, to obtain a large fixed dimensional vector representation, and then to use another \ac{LSTM} (i.e., decoder) to extract the output sequence. However, different from the typical \ac{seq2seq} model, we borrow the idea of the pointer network \citep{Oriol2015} to add a pair-wise attention mechanism to predict the output sequence based on the attention mask over the input sequence. The pair-wise attention is calculated based on an \ac{ASNN} which was previously used for travel mode prediction \citep{Wang2020DeepFunctions}. Model details will be shown in the following sections.

Intuitively, the \ac{LSTM} encoder and decoder aim to capture the global view of the input information (i.e., overall sequence pattern) by embedding the input sequence to hidden vector representation. While the \ac{ASNN}-based pair-wise attention aims to capture the local view (i.e., the relationship between two stops). Our experiments in Section \ref{sec_case} demonstrate the importance of both global and local views in the sequence prediction.

\subsubsection{\ac{LSTM} encoder.}
Given an arbitrary stop sequence $(s_1,...,s_n)$ as the input, let $x_i \in \mathbb{R}^{K}$ be the features of stop $s_i$, where $x_i$ may include the package information, the customer information, and the geographical information of the stop $s_i$. $K$ is the number of features. The encoder computes a sequence of encoder output vectors $(e_1, ..., e_n)$ by iterating the following:
\begin{align}
    h_i^{\text{E}}, e_i &= \text{LSTM}(x_i, h_{i-1}^{\text{E}}; \; \theta^{\text{E}} ) \quad \forall i = 1,...,n
\end{align}
where $h_i^{\text{E}} \in \mathbb{R}^{K_h^\text{E}}$ is the encoder hidden vector with $h_0^{\text{E}} := 0$. $e_i \in \mathbb{R}^{K_e}$ is the encoder output vector. ${K_h^\text{E}}$ and $K_e$ are corresponding vector dimensions. $\theta^{\text{E}}$ is the learnable parameters in an encoder \ac{LSTM} cell. The calculation details of an \ac{LSTM} cell can be found in \ref{app_lstm}. The encoding process transforms a sequence of features $(x_1,...,x_n)$ into a sequence of embedded representation $(e_1,...,e_n)$. And the hidden vector of the last time step ($h_n^{\text{E}}$) includes the global information of the whole sequence, which will be used for the \ac{LSTM} decoder.

\begin{figure}[htb]
\centering
{\includegraphics*[width = 0.6 \linewidth]{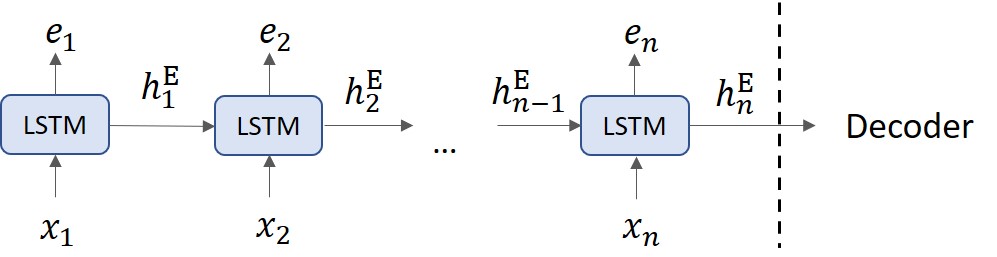}}
\caption{Illustration of \ac{LSTM} ecnoder\label{fig_lstm_e}}
\end{figure}

\subsubsection{\ac{LSTM} decoder.}
\begin{figure}[htb]
\centering
{\includegraphics*[width = 1 \linewidth]{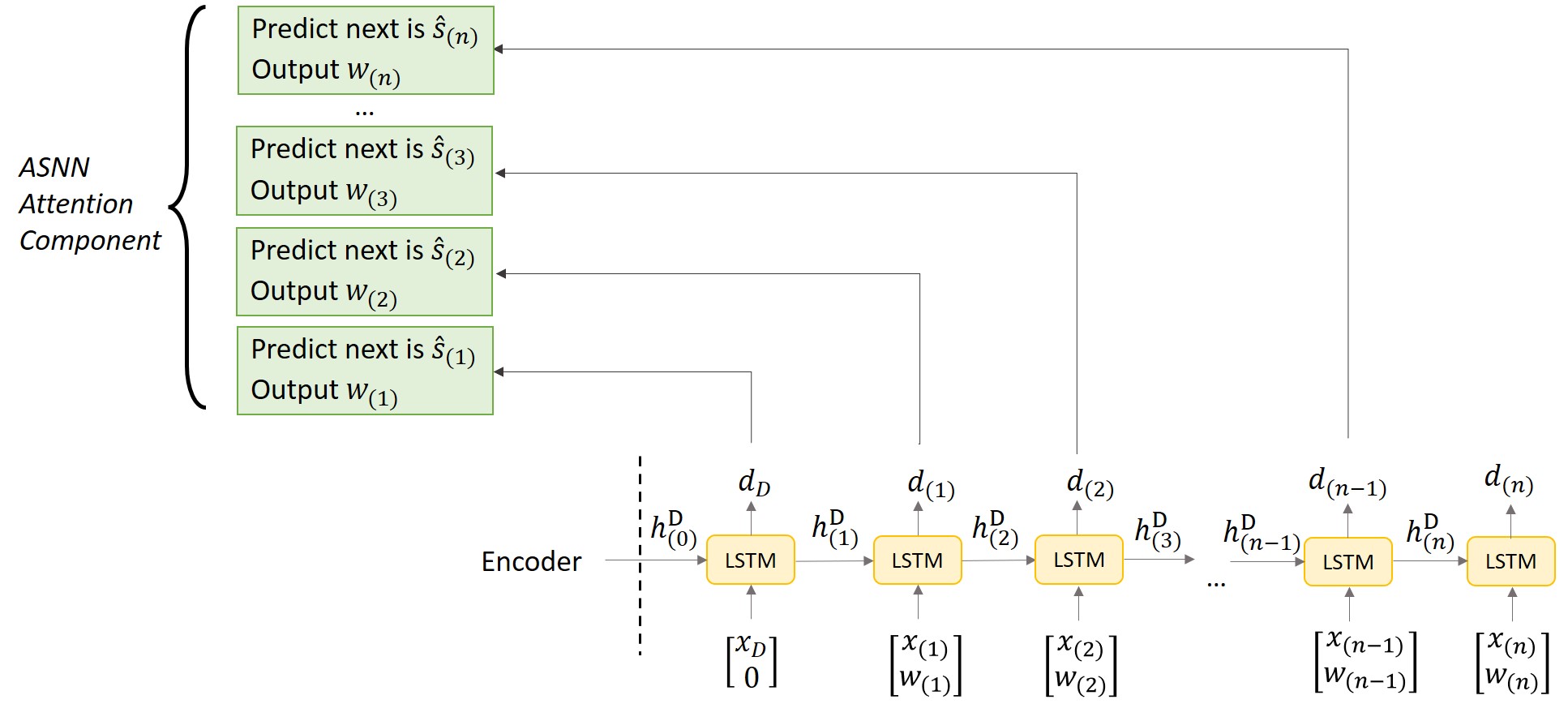}}
\caption{Illustration of \ac{LSTM} decoder\label{fig_lstm_d}}
\end{figure}

The role of a decoder in the traditional \ac{seq2seq} model (Figure \ref{fig_lstm_d}) is to predict a new sequence one time step at a time. However, in the pointer network structure with attention, the role of the decoder becomes producing a vector to modulate the pair-wise attention over inputs.  Denote the output sequence as $(\hat{s}_{(1)}, ..., \hat{s}_{(n)})$. Let $x_{(i)}$ be the feature of stop $\hat{s}_{(i)}$.

At decoder step $i$, we have
\begin{align}
    h_{(i+1)}^{\text{D}}, d_{(i)} &= \text{LSTM}\left(\begin{bmatrix}
x_{(i)} \\
w_{(i)}
\end{bmatrix}, h_{(i)}^{\text{D}} ; \;\theta^{\text{D}}\right) \quad \forall i = 0,1,...,n
\end{align}
where $h_{(i)}^{\text{D}} \in \mathbb{R}^{K_h^\text{D}}$ is the decoder hidden vector with $h_{(0)}^{\text{D}} = h_n^{\text{E}}$, $d_{(i)} \in \mathbb{R}^{K_d}$ is the decoder output vector, ${K_h^\text{D}}$ and $K_d$ are corresponding vector dimensions, and $\theta^{\text{D}}$ are learnable parameters of the decoder \ac{LSTM} cell. Note that we set $x_{(0)} = x_D$ and $d_{(0)} = d_D$, representing the features and the decoder output of the depot, respectively. $w_{(i)}$ is the context vector calculated from the attention component, which will be explained in the next section.

\subsubsection{\ac{ASNN}-based pair-wise attention.}
The pair-wise attention aims to aggregate the global and local information to predict the next stop. Specifically, at each decoder time step $i \in \{0,...,n\}$, we know that the last predicted stop is $\hat{s}_{(i)}$. To predict $\hat{s}_{(i+1)}$, we consider all candidate stops $s_j \in \mathcal{S}$, which is the set of all stops not yet visited. We want to evaluate how possible that $s_j$ will be the next stop of $\hat{s}_{(i)}$. The information of the stop pair $\hat{s}_{(i)}$ and $s_j$ can be represented by the following concatenated vector:
\begin{align}
    v_{(i)}^j = \text{concat}(z_{(i)}^j, \; \phi(x_{(i)}, x_j), \; d_{(i)}, \; e_j)
\end{align}
where $z_{(i)}^j$ is a vector of features associated with the stop pair (such as travel time from $\hat{s}_{(i)}$ to $s_j$), and $\phi(x_{(i)}, x_j)$ represents a feature processing function to extract the pair-wise information from $x_{(i)}$ and $x_j$. For example, $\phi(\cdot)$ may return geographical relationship between stops $\hat{s}_{(i)}$ and $s_j$, and it may also drop features not useful for the attention calculation. Intuitively, $z_{(i)}^j$ and $\phi(x_{(i)}, x_j)$ contains only local information of the stop pair, while  $ d_{(i)}$ and $e_j$ contain the global information of the whole stop set and previously visited stops. 

\begin{figure}[htb]
\centering
{\includegraphics*[width = 0.7 \linewidth]{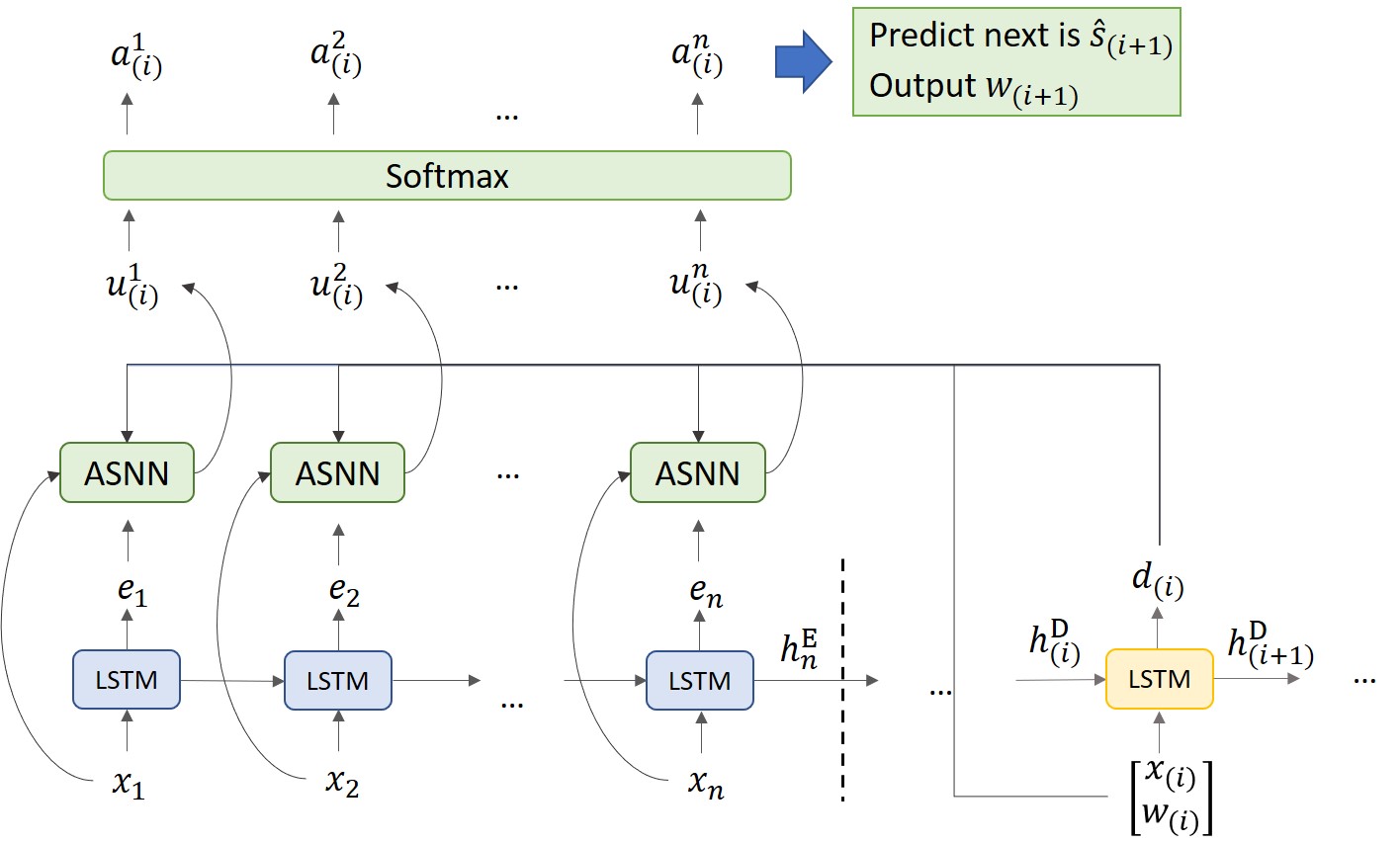}}
\caption{Illustration of \ac{ASNN}-based pair-wise attention\label{fig_Attention}}
\end{figure}

Given the pair-wise information vector $v_{(i)}^j$, we can calculate the attention of stop $\hat{s}_{(i)}$ to stop ${s}_{j}$ as:
\begin{align}
    u_{(i)}^j &= \text{ASNN}(v_{(i)}^j; \; \theta^{\text{A}}) \quad \forall i,j = 1,...,n \label{eq_att_asnn}\\
    a_{(i)}^j &= \frac{\exp({u_{(i)}^j}) }{\sum_{j'=1}^n \exp({u_{(i)}^{j'}}) } \quad \forall i,j = 1,...,n
\end{align}
where $a_{(i)}^j \in \mathbb{R}$ is attention of stop $\hat{s}_{(i)}$ to stop ${s}_{j}$. $\text{ASNN}(\cdot; \;  \theta^{\text{A}}))$ is a multilayer perception (MLP) with the output dimension of one (i.e., $ u_{(i)}^j \in \mathbb{R}$). $\theta^{\text{A}}$ are the learnable parameters of the \ac{ASNN}. The name ``alternative specific'' is because the same parametric network will be applied on all alternative stops $s_j \in \mathcal{S}$ separately \citep{Wang2020DeepFunctions}. Finally, we calculate the conditional probability to make the prediction:
\begin{align}
    &\mathbb{P}(c_{i+1} = j \mid c_1,...,c_{i}, {\mathcal{S}}, X^{\mathcal{S}}; \theta) = a_{(i)}^j \quad \forall i = 0,1,...,n, \;j = 1,...,n \label{eq_att1} \\
    &\hat{s}_{(i+1)} = \argmax_{s_j \in \mathcal{S} \setminus \mathcal{S}_{(i)}^{\text{V}} } a_{(i)}^j \quad \forall i =0, 1,...,n\label{eq_att2}
\end{align}
where $\mathcal{S}_{(i)}^{\text{V}} = \{\hat{s}_{(1)},...,\hat{s}_{(i)}\}$ is the set of stops that have been predicted (i.e., previously visited) until decoder step $i$. Eqs. \ref{eq_att1} and \ref{eq_att2} indicate that the predicted next stop at step $i$ is the one with highest attention among all stops that have not been visited. 

The pair-wise attention framework also leverages the attention information as the input for the next step. This was achieved by introducing the context vector \citep{Bahdanau2015NeuralTranslate}:
\begin{align}
w_{(i)} = \sum_{j = 1}^n a_{(i)}^j \cdot e_j
\end{align}
The context vector is a weighted sum of all the encoder output vectors with attention as the weights. As the attention provides the emphasis for stop prediction, $w_{(i)}$ helps to incorporate the encoded representation of the last predicted stop for the next stop prediction. The inputs for the next \ac{LSTM} cell thus will be the concatenation of the stop features and $w_{(i)}$, i.e.,   $\begin{bmatrix}
x_{(i)} \\
w_{(i)}
\end{bmatrix}$.

It is worth noting that, the specific architecture of $\text{ASNN}(\cdot; \;  \theta^{\text{A}}))$ can be flexible depending on the input pair-wise information. For example, if the information includes images or networks, convolutional  neural network or graph convolutional networks can be used for better extract features. In this study, we use the MLP for simplification as it already outperforms benchmark models. The key idea is of the \ac{ASNN} is to share the same trainable parameter $\theta^{\text{A}}$ for all stop pairs so as to better capture various pair-wise information in the training process. 

\subsection{Sequence generation and selection}\label{sec_seq_gene}
During inference, given a stop set $\mathcal{S}$, the trained model with learned parameters $\theta^*$ are used to generate the sequence. Typically, in the \ac{seq2seq} modeling framework, the final output sequence is selected as the one with the highest probability, i.e.,
\begin{align}
   (s_{j_1^*},...,s_{j_n^*}), \text{where } j_1^*,...,j_n^* = \argmax_{j_1,...,j_n \in \mathcal{C}^{\mathcal{S}} } \mathbb{P}(c_1 = j_1,...,c_n=j_n \mid {\mathcal{S}}, X^{\mathcal{S}}; \theta^*)
\end{align}
where $\mathcal{C}^{\mathcal{S}} = \{\text{All permutations of $\{1,...,n\}$}\}$

Finding this optimal sequence is computationally impractical because of the combinatorial number of possible output sequences. And so it is usually done with the greedy algorithm (i.e., always select the most possible next stop) or the beam search procedure (i.e., find the best possible sequence among a set of generated sequences given a beam size). However, in this study, we observe that the first predicted stop $\hat{s}_{(1)}$ is critical for the quality of the generated sequence. The reason may be that the local relationship between a stop pair (i.e., given the last stop to predict the next one) is easier to learn than the global relationship (i.e., predict the whole sequence). Hence, in this study, we first generate sequences using the greedy algorithm with different initial stops, and select the one with the lowest operational cost. The intuition behind this process is that, once the first stop is given, the model can follow the learned pair-wise relationship to generate the sequence with relatively high accuracy. For all the generated sequences with different first stops, the one with the lowest operation cost captures the global view of the sequence's quality. Therefore, the final sequence generation and selection algorithm is as follows:

\begin{algorithm}
\small
\caption{Sequence generation}
\begin{algorithmic}[1]
\renewcommand{\algorithmicrequire}{\textbf{Input:}}
\renewcommand{\algorithmicensure}{\textbf{Output:}}
\Require Trained model, $\mathcal{S}$
\Ensure  Predicted stop sequence
\For {$s$ in $\mathcal{S}$}
\State Let the first predicted stop be $\hat{s}_{(1)} = s$
\State Predict the following stop sequence $(\hat{s}_{(2)},...,\hat{s}_{(n)})$ using the greedy algorithm. Denote the predicted sequence as $P_s$.
\State Calculate the total operation cost of the whole sequence (including depot), denoted as $OC_s$.
\EndFor
\Return $P_{s^*}$ where $s^* = \argmin_{s\in\mathcal{S}} OC_s$ 
\end{algorithmic} 
\label{alg_seq_generation}
\end{algorithm}

\section{Case Study}\label{sec_case}
\subsection{Dataset}
The data used in our case study was made available as part of the Amazon Last Mile Routing Research Challenge \citep{merchan20222021}. The dataset contains a total of 6,112 actual Amazon driver trajectories for the last-mile delivery from 5 major cities in the US: Austin, Boston, Chicago, Los Angeles, and Seattle. Each route consists of a sequence of stops. Each stop represents the actual parking location of the driver, and the package information (package numbers, package size, and planned service time) associated with each stop is given. The stops are characterized by their latitudes and longitudes, and expected travel time between stops are known.

\begin{figure}[htbp]
\centering
\subfloat[Number of stops distribution]{\includegraphics[width=0.7\textwidth]{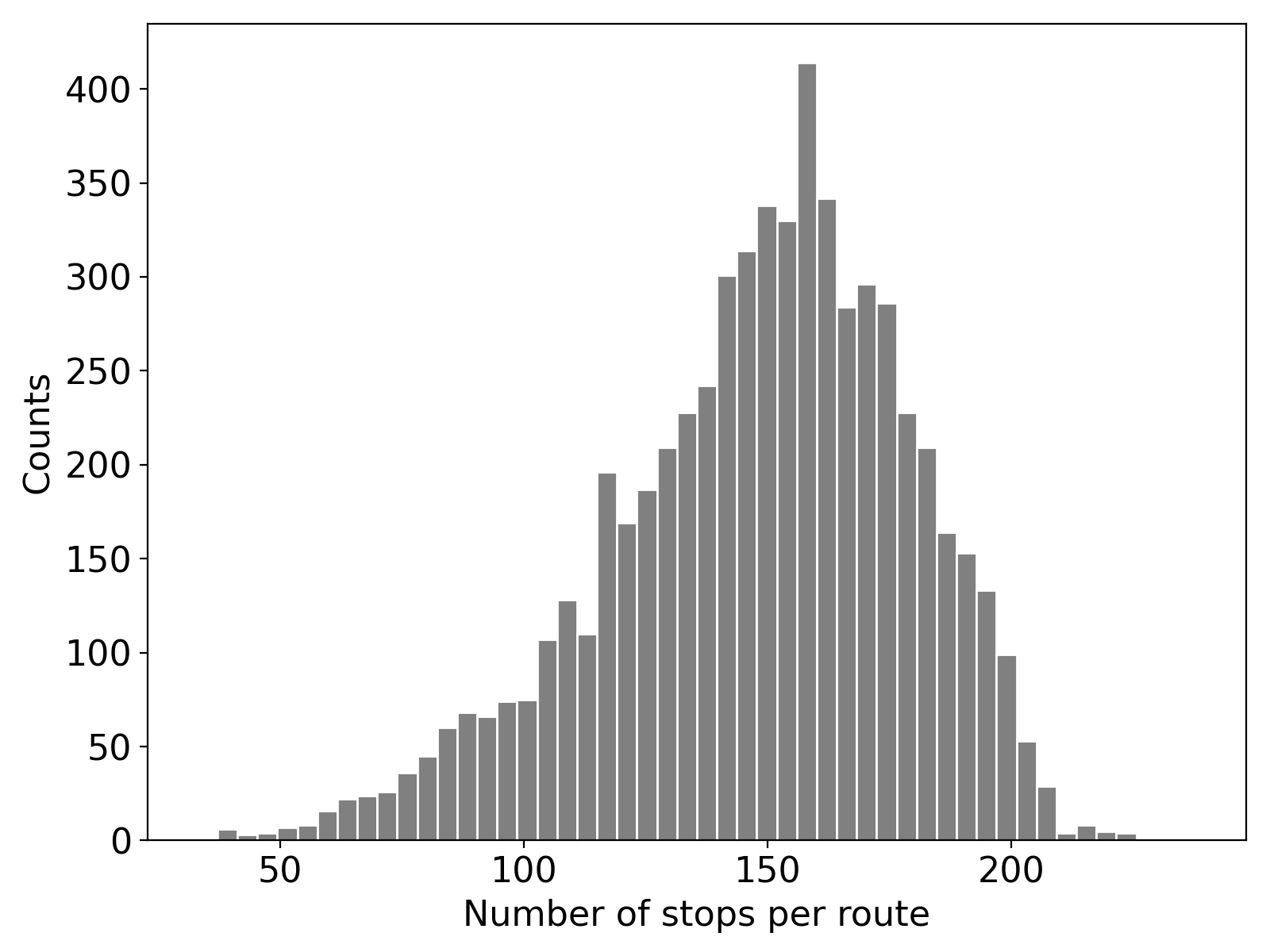}\label{fig_stop_distribution}}
\hfil
\subfloat[Actual route example]{\includegraphics[width=0.7\textwidth]{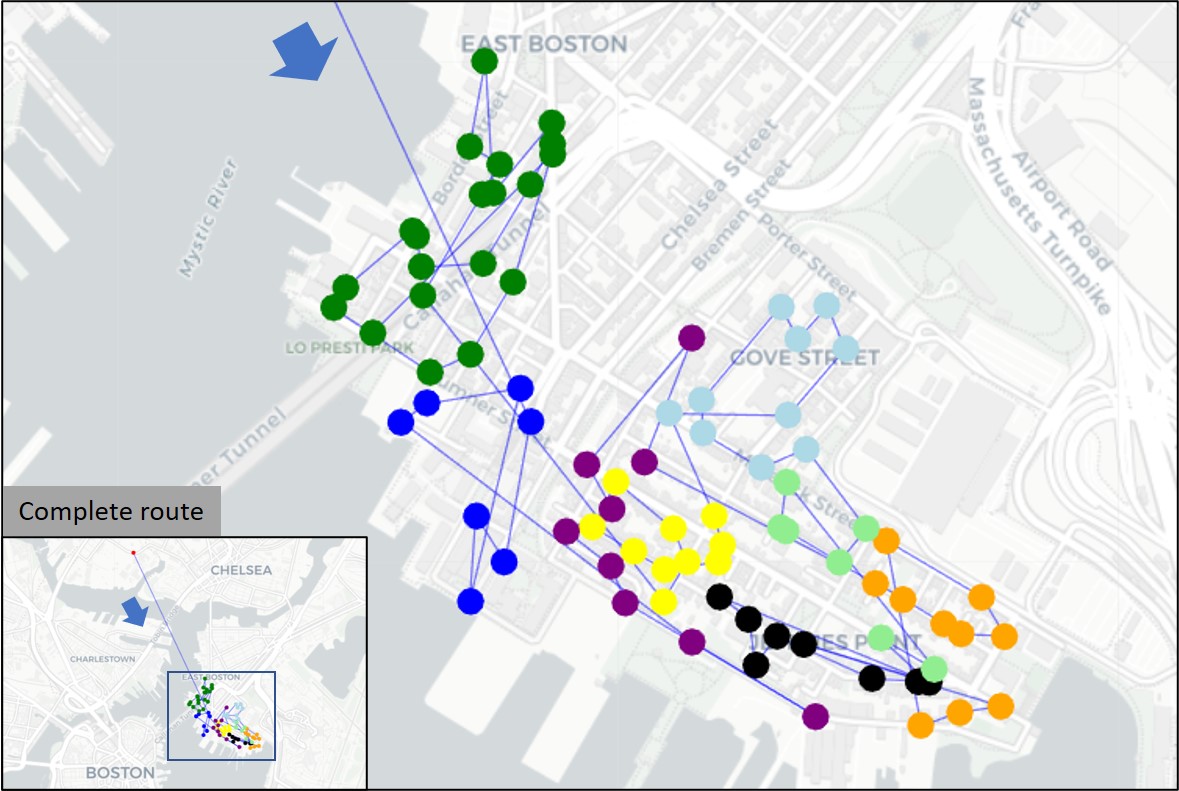}\label{fig_route_example}}
\caption{Description of dataset}
\label{fig_dataset}
\end{figure}

Figure \ref{fig_dataset} shows the distribution of the number of stops per route and an example route. Most routes have around 120 to 180 stops, and the maximum observed number of stops is around 250. Figure \ref{fig_route_example} shows an example of an actual driver trajectory in Boston. Since the depot is far from the delivery stops, we attach the complete route (with the depot indicated by a red dot) at the bottom left of the figure, while the main plot only shows the delivery stops.

\begin{figure}[H]
    \centering
    \includegraphics[width = 0.8 \linewidth]{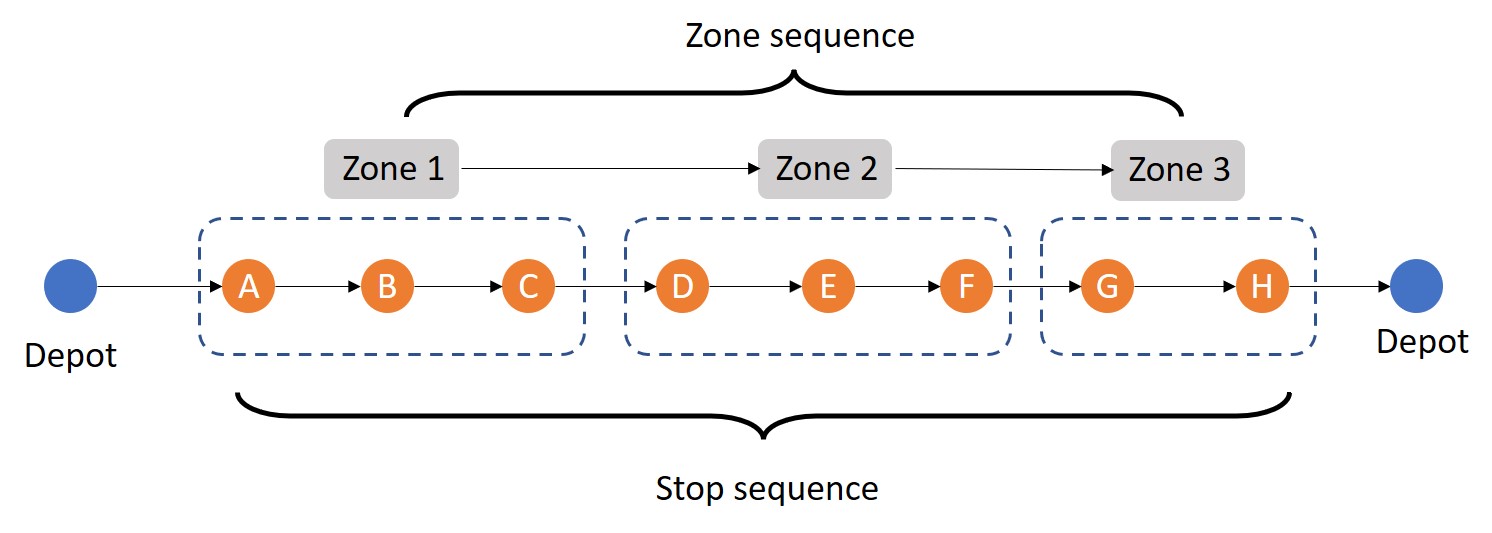}
    \caption{Relationship between stop sequence and zone sequence.}
    \label{fig_zone_seq}
\end{figure}

In this data set, each stop is associated with a zone ID (indicated by different colors in Figure \ref{fig_route_example}). When Amazon generates planned routes for drivers, they usually expect drivers to finish the delivery for one zone first, then go to another zone. And the actual driver trajectories also follow this pattern as shown in Figure \ref{fig_route_example} (but the actual zone sequence may be different from the planned one). Therefore, in this study, we focus on the problem of \emph{zone sequence prediction}. That is, $s_i$ in the case study section now represents a specific zone, $\mathcal{S}$ represents the set of zones, and $X^\mathcal{S}$ represents zone features. This transformation does not affect the model structure proposed in Section \ref{sec_method}. The only difference is that the new problem has a relatively smaller scale compared to the \emph{stop sequence prediction} because the number of zones in a route is smaller than that of stops. The zone-to-zone travel time is calculated as the average travel time of all stop pairs between the two zones. Figure \ref{fig_zone_seq} presents an illustrative example of the relationship between zone and stop sequences. As the dataset does not contain the original planned sequence, we assume the planned zone sequence is the one with the lowest total travel time (generated by a \ac{TSP} solver, $(s_1^{\text{T}}, ..., s_n^{\text{T}})$). After generating the zone sequence, we can restore the whole stop sequence by assuming that drivers within a specific zone follow an optimal \ac{TSP} tour. Details of the zone sequence to stop sequence generation can be found in \ref{app_within_zone_tsp}. 

\subsection{Experimental setup}\label{sec_exp_setup}

We randomly select 4,889 routes for model training and cross-validation, and the remaining 1,223 routes are used to evaluate/test model performance. 

We consider a one-layer \ac{LSTM} for both the encoder and decoder with the hidden unit sizes of 32 (i.e., $K_h^\text{D} = K_e = K_h^\text{E} = K_d = 32$). And the \ac{ASNN} is set with 2 hidden layers with 128 hidden units in each layer. We train the model using Adam optimizer with a default learning rate of 0.001 and 30 training epochs. To utilize the planned route information, the input zone sequence for the \ac{LSTM} encoder is set as the \ac{TSP} result (i.e., lowest travel time). That is, the input sequence $(s_1,...,s_n) = (s_1^{\text{T}}, ..., s_n^{\text{T}})$.

In the case study, $x_i$ represents zone features, including the latitude and longitude of the zone center, number of stops in the zone, number of intersections in the zone, number of packages in the zone, total service time in the zone, total package size in the zone, and the travel time from this zone to all other zones. The zone pair features $z_{(i)}^j$ includes the travel time from $\hat{s}_{(i)}$ to ${s}_{j}$ and zone ID relationship characteristics. For example, the zone IDs ``B-6.2C'' and ``B-6.3A'' signal that they belong to the higher-level cluster ``B-6''. As we assume all pair-wise features are captured by $z_{(i)}^j$, $\phi(x_{(i)},x_j)$ is not specified in this case study.  

\subsection{Evaluation Metrics}
\subsubsection{Disparity score}
Consistent with the Amazon Last Mile Routing Research Challenge, we evaluate the quality of the predicted stop sequences using a ``disparity score'' defined as follows:
\begin{align}
    R(A,B) = \frac{SD(A,B)\cdot ERP_{\text{norm}}(A,B)}{ERP_{\text{e}}(A,B)}
\end{align}
where $R(A,B)$ is the disparity score for the actual sequence $A$ and predicted sequence $B$, and $SD(A,B)$ is the sequence deviation defined as
\begin{align}
    SD(A,B) = \frac{2}{n(n-1)}\sum_{i = 2}^n\left(|c_{[B_i]} - c_{[B_{i-1}]}| -1\right)
\end{align}
where $n$ is the total number of stops, $B_i$ is the $i$-th stop of sequence $B$, $c_{[B_i]}$ is the index of stop $B_i$ in the actual sequence $A$ (i.e., its position in sequence $A$). In the case of $A = B$ (i.e., perfectly predicted), we have  $c_{[B_i]} - c_{[B_{i-1}]} = 1$ for all $i = 2,...,n$, and $SD(A,B) = 0$. 

$ERP_{\text{norm}}(A,B)$ is the Edited Distance with Real Penalty (ERP) defined by the following recursive formula: 
\begin{align}
ERP_{\text{norm}}(A,B)= ERP_{\text{norm}}(A_{2:|A|},B_{2:|B|}) +  \textsc{Time}_{\text{norm}}(A_1, B_1) 
\end{align}

where $\textsc{Time}_{\text{norm}}(s_i,s_j) = \frac{\textsc{Time}(s_i,s_j)}{\sum_{j' \in \{1,...,n\}}\textsc{Time}(s_i,s_{j'})}$ is the normalized travel time from stop $s_i$ to stop $s_j$. $ERP_{\text{e}}(A,B)$ is the number of edit operations (insertions, substitutions, or deletions) required to transform sequence $A$ to sequence $B$ as when executing the recursive $ERP_{\text{norm}}$ formulation. Hence, the ratio $\frac{ERP_{\text{norm}}(A,B)}{ERP_{\text{e}}(A,B)}$ represents the average normalized travel time between the two stops involved in each ERP edit operation.  In the case of $A = B$, we have $\frac{ERP_{\text{norm}}(A,B)}{ERP_{\text{e}}(A,B)} = 0$. 

The disparity score $R(A,B)$ describes how well the model-estimated sequence matches the known actual sequence. Lower score indicates better model performance. A score of zero means perfect prediction. The final model performance is evaluated by the mean score over all routes in the test set. 

\subsubsection{Prediction accuracy}
In addition to the disparity score, we also evaluate the prediction accuracy of the first four zones in each route. We choose the first four zones because the minimum number of zones in a route is four. Let the predicted sequence of the $m$-th route be $A^{(m)}$ and the associated actual sequence be $B^{(m)}$. The prediction accuracy of the $i$-th zone is defined as:
\begin{align}
    \text{Prediction accuracy}_{i} = \frac{\sum_{m=1}^M \mathbbm{1}_{\{A^{(m)}_{i} = B^{(m)}_{i}\} }}{M}
\end{align}
where $M$ is the total number of testing samples. $\mathbbm{1}_{\{\cdot\}}$ is an indicator function which returns 1 if the condition is true, otherwise 0.

\subsection{Benchmark models}

The following optimization and machine learning models are used as benchmarks to compare with the proposed approach.

\textbf{Tour \ac{TSP}}. 
The first benchmark model is the zone sequence generated by tour \ac{TSP}. The first and last stops are both set as the depot. This is the route that we treat as the planned route with the lowest travel time. Since a driver would follow the ``planned routes'' in most of the time, the output from a TSP can be treated as a prediction to the driver's actual route. Note that we cannot use the same objective function in Eq. \ref{eq_obj_LL} for TSP because TSP model cannot evaluate the likelihood of a route. 

\textbf{Open-tour \ac{TSP}}. 
Another variant for TSP models is the open-tour TSP, where we assume drivers do not need to return to the depot. The intuition for this model is that some drivers, when delivering packages, may ignore the last trip back to the depot in their routing decisions. 

\textbf{\ac{ASNN} model}.
The \ac{ASNN} component can be trained to predict the next zone given the current zone, and the prediction sequence can be constructed in a greedy way starting from the given depot. The training zone pairs (including from depot to the first zone) are extracted from all sequences in the training routes. And the input features are the same as the \ac{ASNN} component in the proposed model except for $(d_{(i)}, e_j)$ (i.e., output vectors from \ac{LSTM} decoder and encoder, respectively). All hyper-parameters of the \ac{ASNN} model are the same as the attention component. 

Inspired by the importance of the first zone, we also implement another sequence generation method similar to Section \ref{sec_seq_gene}. That is, we go through all zones in a route and assume it is the first zone, then use the trained \ac{ASNN} to predict the remaining sequence. The final sequence is selected as the one with the lowest travel time. 

\begin{figure}[H]
    \centering
    \includegraphics[width = 0.6 \linewidth]{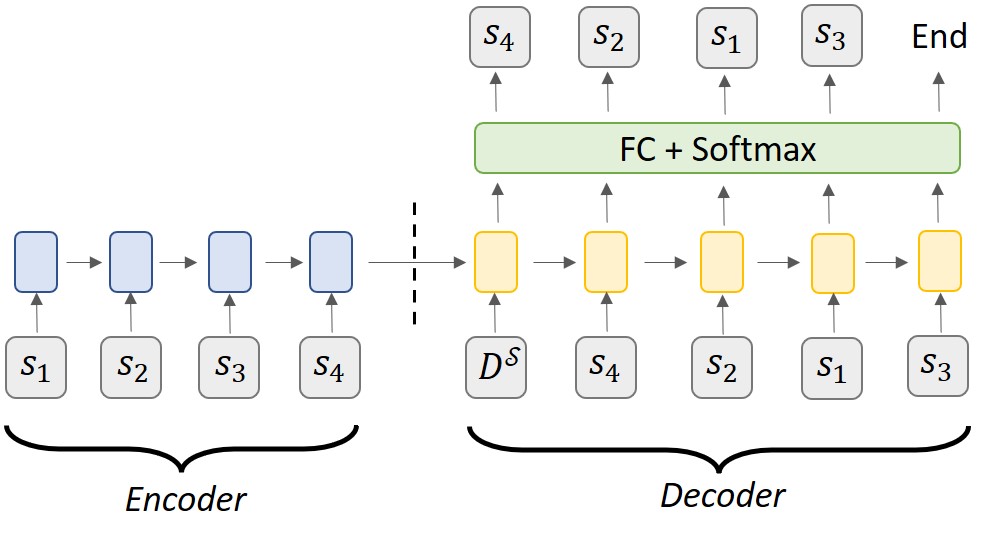}
    \caption{Model architecture of the \ac{LSTM-E-D} \ac{seq2seq} prediction model.}
    \label{fig_lstm_en_de}
\end{figure}

\textbf{\acl{LSTM-E-D}}.
The \acf{LSTM-E-D} architecture is a typical \ac{seq2seq} model proposed by \citet{Sutskever2014SequenceNetworks}. The model structure is shown in Figure \ref{fig_lstm_en_de}. In the decoder stage, the model outputs the predicted zone based on last predicted zone's information. The model formulation can be written as
\begin{align}
    h_i^{\text{E}}, e_i &= \text{LSTM}(x_i, h_{i-1}^{\text{E}}; \; \theta^{\text{E}} ) \quad \forall i = 1,...,n \\ 
      h_{(i+1)}^{\text{D}}, d_{(i)} &= \text{LSTM} (x_{(i)}, h_{(i)}^{\text{D}} ; \;\theta^{\text{D}}) \quad \forall i = 0,1,...,n
\end{align}
The decoder output vector $d_{(i)}$ are, then feed into a fully-connected (FC) layer to calculate probability of the next stop:
\begin{align}
    & g_{(i)} = \text{FC}(d_{(i)} ;\; \theta^{\text{FC}} ) \quad \forall i = 1,...,n \\ 
    & \mathbb{P}(c_{i+1} \mid c_1,...,c_{i}, {\mathcal{S}}, X^{\mathcal{S}}; \theta) = \text{Softmax}(g_{(i)}) \quad \forall i = 1,...,n 
\end{align}
where $g_{(i)} \in \mathbb{R}^{K_z}$, $K_z$ is the maximum number of zones in the dataset. And the next predicted stop is selected by maximizing $\mathbb{P}(c_{i+1} = j \mid c_1,...,c_{i}, {\mathcal{S}}, X^{\mathcal{S}}; \theta)$ for all $s_j \in \mathcal{S} \setminus \mathcal{S}_{(i)}^{\text{V}} $ (i.e., the zones that are not in the route and that have been visited are excluded).

\textbf{Original Pointer Network}.
Another benchmark model is the original pointer network (Pnt Net) proposed by \citet{Oriol2015}. The overall architecture of the pointer network is similar to the proposed model except for the attention component. Specifically, the pointer network calculates attention as:
\begin{align}
    u_{(i)}^j &= W_1^T\text{tanh}(W_2 e_j + W_3 d_{(i)}) \quad \forall i,j = 1,...,n \label{eq_pt_att}\\
    a_{(i)}^j &= \frac{\exp({u_{(i)}^j}) }{\sum_{j'=1}^n \exp({u_{(i)}^{j'}}) } \quad \forall i,j = 1,...,n
\end{align}
The original pointer network does not include the pair-wise local information $(z_{(i)}^j, \; \phi(x_{(i)}, x_j))$, and the attention calculation is only quantified from three learnable parameters $W_1, W_2$, and $W_3$, which may limit its capacity in prediction. We observe that the original pointer network without local information performs extremely badly. For a fair comparison, we add the local information with the similar format in Eq. \ref{eq_pt_att} as:
\begin{align}
    u_{(i)}^j &= W_1^T\text{tanh}(W_2 e_j + W_3 d_{(i)}) + W_4\begin{bmatrix}
z_{(i)}^j \\
\phi(x_{(i)}, x_j)
\end{bmatrix} \quad \forall i,j = 1,...,n 
\label{eq_pt_att2}
\end{align}
After training the model, we generate the final sequence with the greedy algorithm and Algorithm \ref{alg_seq_generation}, respectively.

\textbf{Amazon Last-Mile Routing Research Challenge (ALMRRC) Winning Teams Solutions}. As the paper's data set comes from the ALMRRC, we also compare our models with the top 3 winning teams in the competition. All top 3 winning teams use TSP-based methods. Specifically, \citet{cook2022constrained} proposed a penalty-based local search algorithm for route optimization in the presence of a series predefined constraints, including warehouse sorting operations, van-loading processes, and driver preferences (learned from the data). \citet{guo2023amazon} proposed a hierarchical TSP optimization model with a customized cost matrix. The higher level TSP solves for the zone sequence while the lower level TSP solves the intra-zonal stop sequence. The cost matrix is modified to account for warehouse operations, geographical properties, package information, etc. Similarly, \citet{arslan2021data} used a genetic algorithm to solve a TSP with modified cost matrix and transformed network. The cost matrix are transformed network are generated based on descriptive analysis of the data. The disparity scores of the three teams are shown in Table \ref{tab_results}. 

\subsection{Results}

\subsubsection{Model comparison.}
Table \ref{tab_results} presents the performance of different models. Note that for all approaches except for the \ac{TSP}, we generate sequences based on two different methods (greedy and Algorithm \ref{alg_seq_generation}) for comparison. TSP cannot be incorporated with Algorithm \ref{alg_seq_generation} because the solution of the TSP is the sequence with the minimal operation cost. Even if we iterate all stops to be the first stop, the final selection (based on minimal operation cost) will still be the TSP solution. The standard deviation of disparity scores is taken over all testing routes.  

Results show that sequence generation with Algorithm \ref{alg_seq_generation} (i.e., iterating different first zones) can consistently reduce the disparity score for all machine learning methods. It implies that the first zone prediction and the global view (i.e., shortest path) are important for estimating the driver's trajectory. In \ref{appendix_tsp_based}, we also evaluate sequence generation methods by fixing the first stop as the TSP sequences. Results show that fixing the first stop as the tour TSP sequence  slightly decrease the model's performance as tour TSP does not predict the first stop well. Fixing the first stop as the open-tour TSP has the similar performance as the greedy-based method.

The proposed method outperforms all other models, both in disparity scores and prediction accuracy. This means the proposed pair-wise \ac{ASNN}-based attention (Eq. \ref{eq_att_asnn}) has better performance than the original content-based attention (Eq. \ref{eq_pt_att2}). The comparison between \ac{LSTM-E-D} and Pnt Net models demonstrates the effectiveness of the attention mechanism. All machine learning models except for \ac{LSTM-E-D} can outperform the baseline \ac{TSP} sequence with Algorithm \ref{alg_seq_generation} sequence generation method, suggesting that the hidden trajectory patterns can be learned from the data. Between two TSP models, the open-tour TSP is better than tour TSP. This implies that drivers, when making their decisions on routing, may ignore their last returning trip to the depot. 

Another observation is that, the prediction accuracy and disparity score do not always move in the same direction. For example, the \ac{LSTM-E-D} model with Algorithm \ref{alg_seq_generation} sequence generation, though has lower accuracy, shows a better disparity score. This is because the accuracy metric does not differentiate ``how wrong an erroneous prediction is''. By the definition of disparity score, if a stop is $s_i$ but the prediction is $s_j$, and $s_j$ and $s_i$ are geographically close to each other, the score does not worsen too much. This suggests a future research direction in using disparity score as the loss function (e.g., training by \ac{RL}) instead of cross-entropy loss.

In terms of the comparison with ALMRRC winning team solutions, the performance of the proposed method is similar to that of the second-place winning team, but worse than the first-place team who predicts the route with a constrained local search. The reason may be that the local-search rules defined in \citet{cook2022constrained} have incorporated driver's preferences and warehouse operations. And these rules are difficult to be learned by machine learning algorithm if we do not encode them well. Future studies may follow the local-search rules in \citet{cook2022constrained} to design better information extraction mechanisms to improve the model's performance.  

\begin{table}[htbp]
\centering
\caption{Model performance}\label{tab_results}
\resizebox{\textwidth}{!}{
\begin{tabular}{@{}cccccccc@{}}
\toprule
\multirow{2}{*}{Sequence generation} & \multirow{2}{*}{Model} & \multicolumn{2}{c}{Disparity score} & \multicolumn{4}{c}{Prediction accuracy}   \\ \cmidrule(l){3-8} 
                                &                        & Mean            & Std. Dev          & 1st zone & 2nd zone & 3rd zone & 4th zone \\ \midrule
-                               & Tour TSP                    & 0.0443          &  0.0289            &  0.207    &   0.185       &  0.163        &  0.168  \\
-                               & Open-tour TSP                    & 0.0430          &  0.0302            &  0.270    &   0.244       &  0.227        &  0.232 
\\ \midrule
\multirow{4}{*}{Greedy}         & ASNN                   & 0.0470          &  0.0289         &     0.150     &   0.141     &   0.119       &   0.123   \\
                                & LSTM-E-D               & 0.0503          &  0.0313           &   0.207   &   0.183      &  0.161      &   0.166    \\
                                & Pnt Net                & 0.0460         &  0.0309               &  0.224         & 0.204      & 0.186       & 0.165        \\
                                & \textbf{Ours}               & \textbf{0.0417}          &   0.0306                & \textbf{0.241}    & \textbf{0.231}     & \textbf{0.224}   &  \textbf{0.221}       \\ \midrule
\multirow{4}{*}{Algorithm \ref{alg_seq_generation}}    & ASNN                   & 0.0429          &  0.0299                 &  0.221        & 0.213         &   0.203       & 0.195         \\
                                & LSTM-E-D               & 0.0501          & 0.0305             &  0.182    &  0.156       & 0.142         &  0.149     \\
                                & Pnt Net                & 0.0382                &  0.0301                 &  0.286        & 0.273         & 0.262         & 0.274         \\
                                & \textbf{Ours}                  & \textbf{0.0369}          &   0.0301                &  \textbf{0.320}    & \textbf{0.310}       & \textbf{0.303}        & \textbf{0.314}     \\ \midrule
\multirow{3}{*}{\begin{tabular}[c]{@{}c@{}}Amazon Last-Mile  \\Routing Research Challenge \\Winning Teams Solutions\end{tabular}}    & \citet{cook2022constrained}                   & 0.0198          &  N.A.                 &  N.A.        & N.A.        &   N.A.       & N.A.         \\
                                & \citet{guo2023amazon}      & 0.0381          & N.A.             & N.A.    &  N.A.      & N.A.         & N.A.     \\
                                & \citet{arslan2021data}     & 0.0367                &  N.A.                 &  N.A.        & N.A.         & N.A.         & N.A.        \\    \bottomrule      
\end{tabular}
}
\end{table}

\begin{figure}[H]
\centering
{\includegraphics*[width = 0.6 \linewidth]{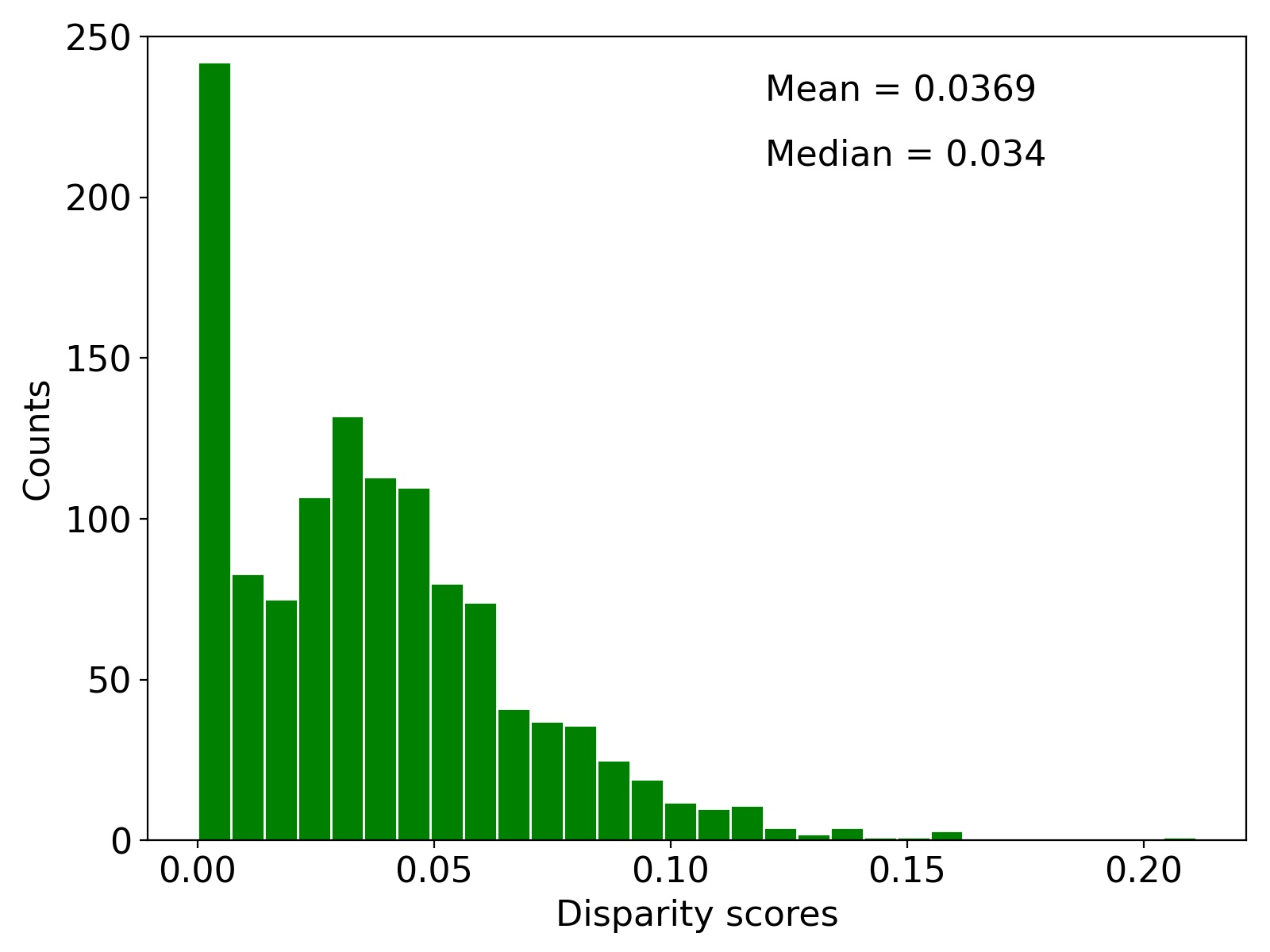}}
\caption{Disparity score distribution of the best model\label{fig_score_distribution}}
\end{figure}

Figure \ref{fig_score_distribution} shows the distribution of disparity scores for our proposed method with Algorithm \ref{alg_seq_generation} sequence generation (i.e., the best model). We observe that the prediction performance varies a lot across different routes. There is a huge proportion of routes with very small disparity scores (less than 0.01). The mean score is impacted by outlier routes. The median score is 0.0340, which is smaller than the mean value.  

\subsubsection{Factors on trajectory predictability}\label{sec_factor}
As our proposed model exhibits various levels of predictability across different routes, we aim to investigate which attributes of a route cause high (or low) predictability. This can be done by running a regression model with the disparity score as the dependent variable and route attributes (e.g., locations, departure time, package numbers) as independent variables. The variables used are defined as follows:
\begin{itemize}
    \item Total planned service time: The estimated time to deliver all packages in the route (service time only, excluding travel time). 
    \item Earliest time window constraint: The earliest due time to deliver packages with time window constraint minus the vehicle departure time. The smaller the value, the tighter the time limit. 
    \item Avg. \# traffic signals: Average number of traffic signals in each zone of the route (obtained from OpenStreetMap data).
    \item If high-quality route: A dummy variable indicating whether the route is labeled as ``high quality'' by Amazon or not (Yes = 1). High quality means the routes have better driver experience, customer satisfaction, and productivity \citep{merchan20222021}.
    \item If in \textsc{Location}: A dummy variable indicating whether the route is in a specific city or not (Yes = 1). 
    \item If departure \textsc{Time}: A dummy variable indicating the (local) departure time (e.g., before 7AM, after 10AM).
\end{itemize}

Table \ref{tab_factor_pred} shows the results of the regression. Since the dependent variable is \emph{disparity scores}, a negative sign indicates a positive impact on the predictability. We observe that routes with tighter time window constraints and more stops are easier to predict. This may be due to the fact that these routes are usually harder to deliver. Hence, to avoid the risk of violating time constraints or delay, drivers tend to follow the planned routes and thus the route sequences are easier predict. We also find that routes associated with larger vans (i.e., larger vehicle capacity) are more predictable. The reason may be that larger vans are less flexible in choosing different routes, thus drivers are more likely to follow the navigation. Another important factor for better predictability is high-quality routes. This may be because high-quality routes are closer to the \ac{TSP} sequence which we use as inputs. Finally, routes in LA are more predictable than in other areas such as Chicago and Boston. 

\begin{table}[htbp]
\centering
\caption{Factors on trajectory predictability}\label{tab_factor_pred}
\begin{tabular}{@{}llllll@{}}
\toprule
Variables & Coefficients ($\times 10^{-3}$) & Variables & Coefficients ($\times 10^{-3}$) \\ \midrule
Intercept          &  91.93 **         &   If high quality route & -5.381 **  \\
Total \# of packages     & 0.079 *        & If in LA  &  -5.312 *           \\ 
Total planned service time    & -0.451             &  If in Chicago &   0.278        \\ 
Earliest time window constraint & -2.970 **             &  If in Boston&   -4.258        \\ 
Avg. \# traffic signals  &  -3.231            &  If on weekends &   1.761    \\ 
Total \# of stops          & -0.171 **             & If departure before 7AM  &  0.617        \\ 
Vehicle capacity ($m^3$)    &  -5.821 *            & If departure after 10AM &  -2.770       \\ 
\bottomrule
\multicolumn{4}{l}{
\begin{tabular}[c]{@{}l@{}} Number of routes: 1,223. \\
$R^2$: 0.072; \\
$^{**}$: $p$-value $<0.01$; $^{*}$: $p$-value $<0.05$.
\end{tabular}} 
\end{tabular}
\end{table}

It is worth noting that the regression model's $R^2$ is relatively small (0.072), implying a low goodness-of-fit. However, the purpose of the regression to identify factors that affect the trajectories' predictability and provides insights on understanding drivers’ behavior. Therefore, it suffices for us to focus on statistically significant parameters (e.g., $p$-value $<$ 0.05), rather than $R^2$.

\subsubsection{Impact of input sequence}
All machine learning models in Table \ref{tab_results} (except for \ac{ASNN}) have the \ac{LSTM} encoder component, which requires the specification of input zone sequence. The input for the encoder is a ``sequence''. The role of encoder is to transforms a sequence of features $(x_1,...,x_n)$ into a sequence of embedded representation $(e_1,...,e_n)$. And the hidden vector of the last time step ($h_n^{\text{E}}$) is expected to include the global information of the whole sequence. Therefore, different input sequences may affect the information extraction in the LSTM encoder component.

As mentioned in Section \ref{sec_exp_setup}, we currently use the \ac{TSP} sequence as input. It is worth exploring the model performance if we use a random zone sequence instead, which corresponds to the scenario without planned route information. Table \ref{tab_result_rand_seq} shows the model performance without the \ac{TSP} sequence information. Since the \ac{ASNN} result does not rely on \ac{TSP} information, it is not listed in the table. Results show that the \ac{LSTM-E-D} model becomes much worse with a random sequence as inputs, while the performance of Pnt Net and our method is only slightly affected. Even without the planned route information, the proposed model can still provide a reasonable estimation of driver trajectories. 

\begin{table}[htbp]
\centering
\caption{Model performance without \ac{TSP} information}\label{tab_result_rand_seq}
\begin{tabular}{@{}cccccccc@{}}
\toprule
\multirow{2}{*}{Sequence generation} & \multirow{2}{*}{Model} & \multicolumn{2}{c}{Disparity score} & \multicolumn{4}{c}{Prediction accuracy}   \\ \cmidrule(l){3-8} 
                                &                        & Mean            & Std. Dev          & 1st zone & 2nd zone & 3rd zone & 4th zone \\ \midrule
\multirow{3}{*}{Greedy}       
                                & LSTM-E-D               & 0.1176          &  0.0498         &     0.045     &   0.047     &   0.041      &   0.050     \\
                                & Pnt Net                & 0.0512         &  0.0323               &  0.090         & 0.096      & 0.097       & 0.096        \\
                                & \textbf{Ours}                    & \textbf{0.0426}          &   0.0311                & \textbf{0.204}    & \textbf{0.192}     & \textbf{0.195}   &  \textbf{0.196}       \\ \midrule
\multirow{3}{*}{Algorithm \ref{alg_seq_generation}}  
                                & LSTM-E-D               & 0.1054          & 0.0463             &  0.103    &  0.061       & 0.049         &  0.052     \\
                                & Pnt Net                & 0.0398                &  0.0311                 &  0.298        & 0.284         & 0.273         & 0.273         \\
                                & \textbf{Ours}                    & \textbf{0.0376}          &   0.0307                &  \textbf{0.316}    & \textbf{0.298}       & \textbf{0.302}        & \textbf{0.298}     \\ \bottomrule
\end{tabular}
\end{table}

\subsubsection{Computational time comparison}
In addition to prediction performance, we also compare the models' computational time. Table \ref{tab_time_comp} shows the computational time of all models in a personal computer of I9-9900K CPU. For the TSP models, we report the solution time for the testing data set of 1,223 routes since it does not require a training process. The models are solved with Gurobi integer optimization solver with subtour elimination as laze constraints. For all machine learning models, we report the model's training time. The evaluation time for generating routes are negligible compared to the training time. 

Results show that the TSP models have the lowest computational time. The reason is that Gurobi solvers with laze constraint implementation is efficient. Another reason is that we only solve for zone sequences. The number of zones is relatively small. In the cases where we have more and longer routes, the computational time for TSP methods can be large (i.e., these methods are not as scalable as machine learning methods). Among all machine learning models, our proposed method, due to its network complexity, has the largest training time. 

\begin{table}[htbp]
\centering
\caption{Computational time comparison}\label{tab_time_comp}
{
{\renewcommand{\arraystretch}{1} 
\begin{tabular}{@{}cccc@{}}
\toprule
Model                 & CPU time (sec)      & Model & CPU time (sec)    \\ \midrule
Open Tour TSP & 49.1  &  Tour TSP & 83.9 \\
LSTM-E-D & 176.0 & Pnt Net     & 397.6   \\         ASNN    & 277.9  & Ours    & 2132.7  \\ \bottomrule
\end{tabular}
}
}
\end{table}

\subsection{Summary}
Our numerical results show that our proposed model outperforms its benchmarks in terms of disparity scores and prediction accuracy, meaning that it can better predict the actual route trajectories taken by drivers. The comparison with benchmark models shows that our proposed \ac{ASNN}-based pair-wise attention mechanism and our sequence generation algorithm (Algorithm \ref{alg_seq_generation}) are both helpful for the prediction. 
Moreover, we can observe that the predictive performance varies across different routes. Factors such as route quality, delivery time windows, and the total number of stops of a route affect predictability. 
Finally, the proposed model is insensitive to the input sequence. The prediction performance only slightly decreases when the input sequence is changed from the \ac{TSP} solution to a random stop sequence. This property implies that we only need the set of stops to implement the model and obtain high-quality solution, while information on the planned route sequence is not strictly required.   

\section{Conclusion and Future Research}\label{sec_dis_con}

In this paper, we propose a pair-wise attention-based pointer neural network that predicts actual driver trajectories on last-mile delivery routes for given sets of delivery stops. Compared to previously proposed pointer networks, this study leverages a new \acl{ASNN}-based attention mechanism to incorporate pair-wise local information (such as relative distances and locations of stops) for the attention calculation. To better capture the global efficiency of a route in terms of operational cost (i.e., total travel time), we further propose a new sequence generation algorithm that finds the lowest-cost route sequence by iterating through different first stops. 

We apply our proposed method to a large set of real operational route data provided by the Amazon Last-Mile Routing Research Challenge in 2021. The results show that our proposed method can outperform a wide range of benchmark models in terms of both the disparity score and prediction accuracy, meaning that the predicted route sequence is closer to the actual sequence executed by drivers. Compared to the best benchmark model (original pointer network), our method reduces the disparity score from 0.0382 to 0.0369, and increases the average prediction accuracy of the first four zones from 0.229 to 0.312.  Moreover, our proposed sequence generation method can consistently improve the prediction performance for all models. The disparity scores are reduced by 10-20\% across different models.
Lastly, we show that the proposed methodology is robust against changes in the input sequence pattern. Compared to an optimal \ac{TSP} solution as the input sequence, a random input sequence only slightly increases the disparity score from 0.0369 to 0.0376.  

The data-driven route planning method proposed in this paper has several highly relevant practical implications. 
First, our proposed model performs well at predicting stop sequences that would be preferable to delivery drivers in a real operational environment, even if it is not provided with a theoretically optimal (i.e., minimal route duration) planned \ac{TSP} sequence as an input. 
Therefore, the model can be used to generate a predicted actual stop sequence that a driver would likely be taking for a given set of delivery stops. The prediction can serve as a new ``empirical'' planned route that is informed by historical driver behavior and thus more consistent with the driver's experience and preferences. 
Second, by comparing the stop sequence predicted by our model with the traditional, \ac{TSP}-based planned stop sequence, a route planner may infer potential reasons for the drivers' deviations and adjust the company's planning procedures and/or driver incentives if necessary.
Third, as stop sequence generation using machine learning models is computationally more efficient than traditional optimization-based approaches, a trained machine learning model can be applied in real-time to quickly re-optimize routes when drivers are unexpectedly forced to deviate from their original stop sequence (e.g., due to road closures) and need updated routing strategies. 

Based on the work presented in this paper, a number of fruitful future research avenues arise.
First, instead of focusing on stop sequence prediction, future work may improve the interpretability of such prediction models and develop machine learning approaches that better explain which factors cause drivers to deviate from a planned stop sequence and how they affect their actual route trajectories \citep{mo2021individual}. 
Second, future work should attempt to combine the strengths of optimization-based route planning approaches and machine learning by incorporating tacit driver knowledge learned via machine learning models into route optimization algorithms.

\appendix
\appendixpage

\section{Mathematical Formulation of a \ac{LSTM} Cell}\label{app_lstm}

The details of an \ac{LSTM} cell, $h_t, e_t = \text{LSTM}(x_t, h_{t-1}; \; \theta)$, is shown below:
\begin{align}
f_t &= \sigma_g(W_{f} x_t + U_{f} h_{t-1} + b_f) \\
i_t &= \sigma_g(W_{i} x_t + U_{i} h_{t-1} + b_i) \\
o_t &= \sigma_g(W_{o} x_t + U_{o} h_{t-1} + b_o) \\
\tilde{c}_t &= \sigma_c(W_{c} x_t + U_{c} h_{t-1} + b_c) \\
c_t &= f_t \circ c_{t-1} + i_t \circ \tilde{c}_t \\
h_t &= o_t \circ \sigma_h(c_t) \\
e_t &= h_t \text{ (if this is a single layer one-directional LSTM)} 
\end{align}
where $[W_f, W_i, W_o, W_c, U_f, U_i, U_o, U_c, b_f, b_i, b_o, b_c] = \theta$ is the vector of learnable parameters. $x_t$ is the input vector to the \ac{LSTM} unit. $f_t$ is the forget gate's activation vector. $i_t$ is the input/update gate's activation vector. $o_t$ is the output gate's activation vector.  $h_t$  is the hidden state vector. $e_t$ is the output vector of the \ac{LSTM}. Note that for a multi-layer or bidirectional \ac{LSTM}, $e_t$ may not equal to $h_t$. In this study, we use a single layer one-directional \ac{LSTM} and thus have $e_t = h_t$.  More details on the output vector can be found in \citet{lstm_pytorch}. $\tilde{c}_t $ is the cell input activation vector. $c_t$ is the cell state vector. ``$\circ$'' indicates the component-wise multiplication.

\section{From Zone Sequence to Stop Sequence}\label{app_within_zone_tsp}

The complete stop sequence is generated based on the given zone sequence. The detailed generation process is shown in Algorithm \ref{alg_inner_seq_generation}.

\begin{algorithm}[!h]
\caption{Complete sequence generation. Input: zone sequence $(\hat{z}_{(1)},..,\hat{z}_{(n)})$, depot $D^{\mathcal{S}}$, set of stops in each zone $\mathcal{S}_{(i)}, i=1,...,n$. $\sfunction{PathTSP}(\mathcal{S}, s_{first}, s_{last})$ and $\sfunction{TourTSP}(\mathcal{S})$ are two oracle functions for solving path and tour \ac{TSP} problems given the set of stops $\mathcal{S}$, first stop $s_{first}$ and last stop $s_{last}$ to be visited.}
\label{alg_inner_seq_generation}
\begin{algorithmic}[1]
\Function{CompleteSeqGeneration}{$(\hat{z}_{(1)},..,\hat{z}_{(n)}), \{\mathcal{S}_{(i)}, i=1,,,n\}$}
    \State $s_{prev} \gets D^{\mathcal{S}}$ 
    \State $s_{complete}^* \gets (s_{prev})$ \Comment{Initialize the complete stop sequence with depot}
    \For{$i \in \{1,...,n-1\}$}
        \State $\mathcal{S}_{first} \gets$ Set of three stops in $\mathcal{S}_{(i)}$ that are closest to $s_{prev}$
        \State $\mathcal{S}_{last} \gets$ Set of three stops in $\mathcal{S}_{(i)}$ that are closest to all stops in $\mathcal{S}_{i+1}$ on average
        \State $\mathcal{P}_{(i)} \gets \emptyset$ \Comment{Initialize the set of optimal paths in zone $\hat{z}_{(i)}$}
        \For{$s_{first} \in \mathcal{S}_{first}$}
            \For{$s_{last} \in \mathcal{S}_{last}$}
                \If{$s_{first} = s_{last}$}
                    \State $\hat{p}_{\text{temp}}, {t}_{\text{temp}} = \sfunction{TourTSP}(\mathcal{S}_{(i)})$ \Comment{Solve the optimal tour and travel time for zone $\hat{z}_{(i)}$}
                    \State Delete the last edge back to $s_{first}$ in the tour $\hat{p}^{\text{temp}}$. Let the new path and travel time be $\hat{p}_{\text{temp}}'$ and ${t}_{\text{temp}}'$ 
                    \State Add $\hat{p}_{\text{temp}}'$ and ${t}_{\text{temp}}'$ to $\mathcal{P}_{(i)}$
                \Else
                    \State $\hat{p}_{\text{temp}}, {t}_{\text{temp}} = \sfunction{PathTSP}(\mathcal{S}_{(i)}, s_{first}, s_{last})$ \Comment{Solve the optimal path and travel time for zone $i$}
                    \State Add $\hat{p}_{\text{temp}}$ and ${t}_{\text{temp}}$ to $\mathcal{P}_{(i)}$
                \EndIf
            \EndFor
        \EndFor
        \State $\hat{p}_{(i)} \gets$ Path in $\mathcal{P}_{(i)}$ with the minimum travel time 
        \State $s_{complete}^* \gets$ $(s_{complete}^*, \;\hat{p}_{(i)}$) \Comment{Concatenate two sequence}
        \State $s_{prev} \gets$ Last stop of path $\hat{p}_{(i)}$
    \EndFor
    \State $s_{complete}^* \gets (s_{complete}^*,\;D^{\mathcal{S}})$ \Comment{Concatenate the last stop as the depot}
    \State \textbf{return} $s_{complete}^*$
\EndFunction
\end{algorithmic}
\end{algorithm}

Consider an optimal zone sequence, $(\hat{z}_{(1)},..,\hat{z}_{(n)})$, generated from the proposed machine learning method. We can always add the depot before the first and after last zone (i.e., $(D^{\mathcal{S}}, \hat{z}_{(1)},..,\hat{z}_{(n)}, D^{\mathcal{S}})$) and make the whole zone sequence a loop. For each zone $\hat{z}_{(i)}$, we aim to generate a within-zone path $\hat{p}_{(i)}$, and the final stop sequence will be $(D^{\mathcal{S}}, \hat{p}_{(i)},...,\hat{p}_{(n)}, D^{\mathcal{S}})$. 

When generating $\hat{p}_{(i)}$ for zone $\hat{z}_{(i)}$, we assume $\hat{p}_{(i-1)}$ is known (generated from the last step and $\hat{p}_{(0)} = (D^{\mathcal{S}})$). Let the set of all stops in zone $\hat{z}_{(i)}$ be $\mathcal{S}_{(i)}$. We identify three \emph{potential} first stops and last stops of path $\hat{p}_{(i)}$ based on following rules:
\begin{itemize}
    \item Three potential first stops of $\hat{p}_{(i)}$ are the three most closest stops (in travel time) to $\hat{p}_{(i-1)}$'s last stop.
    \item Three potential last stops of $\hat{p}_{(i)}$ are the three most closest stops (in travel time) to all stops in $\mathcal{S}_{(i+1)}$ on average. Note that $\mathcal{S}_{(n+1)} = \{D^{\mathcal{S}}\}$
\end{itemize}

With three potential first stops and last stops, we then solve path \ac{TSP} problems between any first and last stop pair to generate the potential optimal inner zone path with the shortest travel time. In this step, at most nine small-scale path \ac{TSP} problems will be solved since there might be overlapping between the first and the last stops. If the first and the last stops are identical, we solve a tour \ac{TSP} problem and output the path by deleting the last edge which traverses back to the first stop in the tour. 

After having all potential inner zone paths and total path travel time between any first and last stop pair, we keep the path with the minimum travel time as the inner zone sequence, $\hat{p}_{(i)}$. The key assumption we make here about drivers is that they will deliver packages within a zone following a path that minimizes their total travel time. With the optimal inner zone stop sequence of the current zone, we then move to the next visited zone in the optimal zone sequence and repeat the same procedure until we generate the complete stop sequence.

\section{TSP-based first stop for sequence generation}\label{appendix_tsp_based}
In Algorithm \ref{alg_seq_generation}, we propose to generate sequences by iterating different first stops. Results in Table \ref{tab_results} show that the prediction of the first stop is important. In this section, we tested another way of determining the first stop by using the first stop predicted by both tour TSP and open-tour TSP (referred to as TSP-based). Results are shown in Table \ref{tab_result_TSP_based}. 

\begin{table}[htbp]
\centering
\caption{Model performance with \ac{TSP}-based first stop}\label{tab_result_TSP_based}
\begin{tabular}{@{}cccccccc@{}}
\toprule
\multirow{2}{*}{Sequence generation} & \multirow{2}{*}{Model} & \multicolumn{2}{c}{Disparity score} & \multicolumn{4}{c}{Prediction accuracy}   \\ \cmidrule(l){3-8} 
                                &                        & Mean            & Std. Dev          & 1st zone & 2nd zone & 3rd zone & 4th zone \\ \midrule
\multirow{3}{*}{Tour TSP-based}       
                                & LSTM-E-D               & 0.0503          &  0.0313         &     0.207     &   0.183     &   0.161      &   0.166     \\
                                & Pnt Net                & 0.0472         &  0.0315               &  0.207         & 0.191      & 0.177       & 0.150        \\
                                & \textbf{Ours}                    & \textbf{0.0429}          &   0.0309                & \textbf{0.207}    & \textbf{0.203}     & \textbf{0.195}   &  \textbf{0.193}       \\ \midrule
 \multirow{3}{*}{Open-tour TSP-based}       
                                & LSTM-E-D               & 0.0531          &  0.0316         &     0.270     &   0.178     &   0.142      &   0.146     \\
                                & Pnt Net                & 0.0457         &  0.0314               &  0.270         & 0.231      & 0.209       & 0.184        \\
                                & \textbf{Ours}                    & \textbf{0.0419}          &   0.0303                & \textbf{0.270}    & \textbf{0.250}     & \textbf{0.231}   &  \textbf{0.229}       \\                               
                                \bottomrule
\end{tabular}
\end{table}

We observe that using the tour TSP-based first stop for sequence generation slightly decrease all model's performance compared to greedy-based sequence (compared to Table \ref{tab_results}). The reason may be that tour TSP does not predict the first stop well (with prediction accuracy of 0.207), thus affecting the following sequence generation accuracy. The open-tour TSP-based first stop sequence generation shows the similar performance as the greedy-based approach. 

\section{Analysis of high-quality routes}\label{appendix_high_quality}
High-quality routes are those with better driver experience, customer satisfaction, and productivity \citep{merchan20222021}. There are a total of 2,718 high-quality routes out of the 6,112 trajectories (551 out of the 1,223 testing data set). In Section \ref{sec_factor}, we have shown that high-quality routes are easier to predict using machine learning models. To better quantify the differences, we compare the model prediction performance of high-quality and non-high-quality routes (Table \ref{tab_high_quality_comparsion}). On average, the high-quality scores show around 6.4\% lower mean disparity scores.

\begin{table}[htbp]
\centering
\caption{Model performance comparison for high and non-high quality routes}\label{tab_high_quality_comparsion}
\begin{tabular}{@{}cccccccc@{}}
\toprule
\multirow{2}{*}{Route sets (\# routes)} & \multirow{2}{*}{Model} & \multicolumn{2}{c}{Disparity score} &   \\ \cmidrule(l){3-4} 
                                &                        & Mean            & Median         \\ \midrule
\multirow{3}{*}{High-quality (551)}       
                                & LSTM-E-D               & 0.0487          &  0.0447    \\
                                & Pnt Net                & 0.0374         &  0.0347          \\
                                & {Ours}                    & {0.0352}          &   0.0319                  \\ \midrule
\multirow{3}{*}{Non-high-quality (672)}  
                                & LSTM-E-D               & 0.0514          & 0.0477             \\
                                & Pnt Net                & 0.0389           & 0.0341       \\
                                & {Ours}                    & {0.0391}          &   {0.0357}           \\ \bottomrule
\end{tabular}
\end{table}

The better model performance in high-quality routes implies that the model's route generation is more aligned with the high-quality routes. This shows a promising implementation of machine learning methods in route planning. That is, instead of generating planned routes using optimization methods, we can directly generate routes using machine learning models. And it can potentially be better (i.e., higher quality) than the lowest cost routes. However, one thing to notice is that machine learning approaches are a black box. It may be dangerous to fully trust the routes generated by machine learning models. On another hand, since route quality is a post-execution indicator (as it includes information about the driver's and customer's experience), there is no way to pre-quantify the route quality. To implement machine learning methods, we may need real-world A/B testing to validate the model's actual performance, and use guardrail metrics (such as operation cost and detouring rates) to eliminate nonsense routes generated by machine learning models. 

Since we do not have the exact definition of route quality, to better understand high-quality routes, we conduct a logistic regression using the dummy variable ``If high-quality route'' (Yes=1) as the dependent variable to figure out the properties of high-quality routes. All 6,112 routes are used as samples. Results are shown in Table \ref{tab_logistic}. We observe that routes with a higher number of packages, more traffic signals along the route, and larger vehicles are more likely to be high-quality routes. The reason can be that, routes with high package volumes may be more productive for drivers because they are more likely to deliver a bundle of packages together. Routes with more traffic signals may be in urban areas and the driver's local knowledge are more useful with more complex traffic conditions (thus more likely to be high-quality). And routes with larger vehicle capacity are more likely to be assigned to experienced drivers, which are also more likely to be high-quality. Factors that make a route less likely to be high-quality include more stops and locations in Chicago and Boston (as opposed to Austin, LA, and Seattle). The reason may be that these factors all imply higher route complexity.

\begin{table}[htbp]
\centering
\caption{Results of logistic regression for high-quality routes}\label{tab_logistic}
\begin{tabular}{@{}llllll@{}}
\toprule
Variables & Coefficients & Variables & Coefficients\\ \midrule
Intercept          &  -2.11 **         &    If on weekends &   0.099  \\
Total \# of packages     & 0.015 **        & If in LA  &  -0.100           \\ 
Total planned service time    & 0.046             &  If in Chicago &   -0.291 **        \\ 
Earliest time window constraint & 0.025             &  If in Boston&   -5.237 **        \\ 
Avg. \# traffic signals  &  0.371 *           & If departure before 7AM  &  0.069   \\ 
Total \# of stops          & -0.021 **             & If departure after 10AM &  -0.156        \\ 
Vehicle capacity ($m^3$)    &  0.294 **            &      \\ 
\bottomrule
\multicolumn{4}{l}{
\begin{tabular}[c]{@{}l@{}} Number of routes: 6,112. \\
Log-Likelihood: -3901.7, Log-Likelihood Null: -4199.1; Pseudo $R^2$: 0.071\\
$^{**}$: $p$-value $<0.01$; $^{*}$: $p$-value $<0.05$.
\end{tabular}} 
\end{tabular}
\end{table}

\bibliography{mybibfile}

\begin{thebibliography}{59}
\expandafter\ifx\csname natexlab\endcsname\relax\def\natexlab#1{#1}\fi
\providecommand{\url}[1]{\texttt{#1}}
\providecommand{\href}[2]{#2}
\providecommand{\path}[1]{#1}
\providecommand{\DOIprefix}{doi:}
\providecommand{\ArXivprefix}{arXiv:}
\providecommand{\URLprefix}{URL: }
\providecommand{\Pubmedprefix}{pmid:}
\providecommand{\doi}[1]{\href{http://dx.doi.org/#1}{\path{#1}}}
\providecommand{\Pubmed}[1]{\href{pmid:#1}{\path{#1}}}
\providecommand{\bibinfo}[2]{#2}
\ifx\xfnm\relax \def\xfnm[#1]{\unskip,\space#1}\fi
\bibitem[{Applegate et~al.(2006)Applegate, Bixby, Chvatal and
  Cook}]{Applegate2006TheProblem}
\bibinfo{author}{Applegate, D.L.}, \bibinfo{author}{Bixby, R.E.},
  \bibinfo{author}{Chvatal, V.}, \bibinfo{author}{Cook, W.J.},
  \bibinfo{year}{2006}.
\newblock \bibinfo{title}{{The Travelling Salesman Problem}}.
\newblock \bibinfo{publisher}{Princeton University Press}.
\newblock \URLprefix \url{http://www.jstor.org/stable/j.ctt7s8xg.9}.
\bibitem[{Arslan and Abay(2021)}]{arslan2021data}
\bibinfo{author}{Arslan, O.}, \bibinfo{author}{Abay, R.}, \bibinfo{year}{2021}.
\newblock \bibinfo{title}{Data-driven Vehicle Routing in Last Mile Delivery}.
\newblock \bibinfo{publisher}{Bureau de Montreal, Universit{\'e} de Montreal}.
\bibitem[{Bahdanau et~al.(2015)Bahdanau, Cho and
  Bengio}]{Bahdanau2015NeuralTranslate}
\bibinfo{author}{Bahdanau, D.}, \bibinfo{author}{Cho, K.H.},
  \bibinfo{author}{Bengio, Y.}, \bibinfo{year}{2015}.
\newblock \bibinfo{title}{{Neural machine translation by jointly learning to
  align and translate}}, in: \bibinfo{booktitle}{3rd International Conference
  on Learning Representations, ICLR 2015 - Conference Track Proceedings}.
\bibitem[{Bello et~al.(2019)Bello, Pham, Le, Norouzi and Bengio}]{Bello2019}
\bibinfo{author}{Bello, I.}, \bibinfo{author}{Pham, H.}, \bibinfo{author}{Le,
  Q.V.}, \bibinfo{author}{Norouzi, M.}, \bibinfo{author}{Bengio, S.},
  \bibinfo{year}{2019}.
\newblock \bibinfo{title}{{Neural combinatorial optimization with reinforcement
  learning}}.
\newblock \bibinfo{journal}{5th International Conference on Learning
  Representations, ICLR 2017 - Workshop Track Proceedings} ,
  \bibinfo{pages}{1--15}.
\bibitem[{Cheikhrouhou and Khoufi(2021)}]{Cheikhrouhou2021}
\bibinfo{author}{Cheikhrouhou, O.}, \bibinfo{author}{Khoufi, I.},
  \bibinfo{year}{2021}.
\newblock \bibinfo{title}{A comprehensive survey on the multiple traveling
  salesman problem: Applications, approaches and taxonomy}.
\newblock \bibinfo{journal}{Computer Science Review} \bibinfo{volume}{40},
  \bibinfo{pages}{100369}.
\newblock \URLprefix \url{https://doi.org/10.1016/j.cosrev.2021.100369},
  \DOIprefix\doi{10.1016/j.cosrev.2021.100369}.
\bibitem[{Cho et~al.(2014)Cho, {van Merrienboer}, Gulcehre, Bougares, Schwenk
  and Bengio}]{cho2014learn}
\bibinfo{author}{Cho, K.}, \bibinfo{author}{{van Merrienboer}, B.},
  \bibinfo{author}{Gulcehre, C.}, \bibinfo{author}{Bougares, F.},
  \bibinfo{author}{Schwenk, H.}, \bibinfo{author}{Bengio, Y.},
  \bibinfo{year}{2014}.
\newblock \bibinfo{title}{Learning phrase representations using rnn
  encoder-decoder for statistical machine translation}, in:
  \bibinfo{booktitle}{Conference on Empirical Methods in Natural Language
  Processing (EMNLP 2014)}.
\bibitem[{Chorowski et~al.(2015)Chorowski, Bahdanau, Serdyuk, Cho and
  Bengio}]{chorowski2015attention}
\bibinfo{author}{Chorowski, J.}, \bibinfo{author}{Bahdanau, D.},
  \bibinfo{author}{Serdyuk, D.}, \bibinfo{author}{Cho, K.},
  \bibinfo{author}{Bengio, Y.}, \bibinfo{year}{2015}.
\newblock \bibinfo{title}{Attention-based models for speech recognition}.
\newblock \bibinfo{journal}{arXiv preprint arXiv:1506.07503} .
\bibitem[{Chung et~al.(2014)Chung, Gulcehre, Cho and
  Bengio}]{chung2014empirical}
\bibinfo{author}{Chung, J.}, \bibinfo{author}{Gulcehre, C.},
  \bibinfo{author}{Cho, K.}, \bibinfo{author}{Bengio, Y.},
  \bibinfo{year}{2014}.
\newblock \bibinfo{title}{Empirical evaluation of gated recurrent neural
  networks on sequence modeling}.
\newblock \bibinfo{journal}{arXiv preprint arXiv:1412.3555} .
\bibitem[{Cook et~al.(2022)Cook, Held and Helsgaun}]{cook2022constrained}
\bibinfo{author}{Cook, W.}, \bibinfo{author}{Held, S.},
  \bibinfo{author}{Helsgaun, K.}, \bibinfo{year}{2022}.
\newblock \bibinfo{title}{Constrained local search for last-mile routing}.
\newblock \bibinfo{journal}{Transportation Science} .
\bibitem[{Cortes and Suzuki(2021)}]{Cortes2021Last-mileIndustry}
\bibinfo{author}{Cortes, J.D.}, \bibinfo{author}{Suzuki, Y.},
  \bibinfo{year}{2021}.
\newblock \bibinfo{title}{{Last-mile delivery efficiency: en route transloading
  in the parcel delivery industry}}.
\newblock \bibinfo{journal}{International Journal of Production Research}
  \bibinfo{volume}{0}, \bibinfo{pages}{1--18}.
\newblock \URLprefix \url{https://doi.org/00207543.2021.1907628},
  \DOIprefix\doi{10.1080/00207543.2021.1907628}.
\bibitem[{Davendra and Bialic-Davendra(2020)}]{Davendra2020}
\bibinfo{author}{Davendra, D.}, \bibinfo{author}{Bialic-Davendra, M.},
  \bibinfo{year}{2020}.
\newblock \bibinfo{title}{Introductory chapter: Traveling salesman problem - an
  overview}, in: \bibinfo{booktitle}{Novel Trends in the Traveling Salesman
  Problem}. \bibinfo{publisher}{{IntechOpen}}.
\newblock \URLprefix \url{https://doi.org/10.5772/intechopen.94435},
  \DOIprefix\doi{10.5772/intechopen.94435}.
\bibitem[{Graves(2013)}]{Graves2013GeneratingNetworks}
\bibinfo{author}{Graves, A.}, \bibinfo{year}{2013}.
\newblock \bibinfo{title}{{Generating Sequences With Recurrent Neural
  Networks}} \URLprefix \url{http://arxiv.org/abs/1308.0850}.
\bibitem[{Guo et~al.(2023)Guo, Mo and Wang}]{guo2023amazon}
\bibinfo{author}{Guo, X.}, \bibinfo{author}{Mo, B.}, \bibinfo{author}{Wang,
  Q.}, \bibinfo{year}{2023}.
\newblock \bibinfo{title}{Amazon last-mile delivery trajectory prediction using
  hierarchical tsp with customized cost matrix}.
\newblock \bibinfo{journal}{arXiv preprint arXiv:2302.02102} .
\bibitem[{Guo and Samaranayake(2022)}]{GUO2022103691}
\bibinfo{author}{Guo, X.}, \bibinfo{author}{Samaranayake, S.},
  \bibinfo{year}{2022}.
\newblock \bibinfo{title}{Shareability network based decomposition approach for
  solving large-scale single school routing problems}.
\newblock \bibinfo{journal}{Transportation Research Part C: Emerging
  Technologies} \bibinfo{volume}{140}, \bibinfo{pages}{103691}.
\newblock \URLprefix
  \url{https://www.sciencedirect.com/science/article/pii/S0968090X22001322},
  \DOIprefix\doi{https://doi.org/10.1016/j.trc.2022.103691}.
\bibitem[{Halim and Ismail(2017)}]{Halim2017}
\bibinfo{author}{Halim, A.H.}, \bibinfo{author}{Ismail, I.},
  \bibinfo{year}{2017}.
\newblock \bibinfo{title}{Combinatorial optimization: Comparison of heuristic
  algorithms in travelling salesman problem}.
\newblock \bibinfo{journal}{Archives of Computational Methods in Engineering}
  \bibinfo{volume}{26}, \bibinfo{pages}{367--380}.
\newblock \URLprefix \url{https://doi.org/10.1007/s11831-017-9247-y},
  \DOIprefix\doi{10.1007/s11831-017-9247-y}.
\bibitem[{Hochreiter and Schmidhuber(1997)}]{hochreiter1997long}
\bibinfo{author}{Hochreiter, S.}, \bibinfo{author}{Schmidhuber, J.},
  \bibinfo{year}{1997}.
\newblock \bibinfo{title}{{Long Short-Term Memory}}.
\newblock \bibinfo{journal}{Neural Computation} \bibinfo{volume}{9},
  \bibinfo{pages}{1735--1780}.
\newblock \URLprefix \url{https://doi.org/10.1162/neco.1997.9.8.1735},
  \DOIprefix\doi{10.1162/neco.1997.9.8.1735},
  \href{http://arxiv.org/abs/https://direct.mit.edu/neco/article-pdf/9/8/1735/813796/neco.1997.9.8.1735.pdf}{\tt
  arXiv:https://direct.mit.edu/neco/article-pdf/9/8/1735/813796/neco.1997.9.8.1735.pdf}.
\bibitem[{Huang et~al.(2018)Huang, Vaswani, Uszkoreit, Shazeer, Simon,
  Hawthorne, Dai, Hoffman, Dinculescu and Eck}]{huang2018music}
\bibinfo{author}{Huang, C.Z.A.}, \bibinfo{author}{Vaswani, A.},
  \bibinfo{author}{Uszkoreit, J.}, \bibinfo{author}{Shazeer, N.},
  \bibinfo{author}{Simon, I.}, \bibinfo{author}{Hawthorne, C.},
  \bibinfo{author}{Dai, A.M.}, \bibinfo{author}{Hoffman, M.D.},
  \bibinfo{author}{Dinculescu, M.}, \bibinfo{author}{Eck, D.},
  \bibinfo{year}{2018}.
\newblock \bibinfo{title}{Music transformer}.
\newblock \bibinfo{journal}{arXiv preprint arXiv:1809.04281} .
\bibitem[{Jonker and Volgenant(1983)}]{Jonker1983}
\bibinfo{author}{Jonker, R.}, \bibinfo{author}{Volgenant, T.},
  \bibinfo{year}{1983}.
\newblock \bibinfo{title}{Transforming asymmetric into symmetric traveling
  salesman problems}.
\newblock \bibinfo{journal}{Operations Research Letters} \bibinfo{volume}{2},
  \bibinfo{pages}{161--163}.
\newblock \URLprefix \url{https://doi.org/10.1016/0167-6377(83)90048-2},
  \DOIprefix\doi{10.1016/0167-6377(83)90048-2}.
\bibitem[{Joshi et~al.(2019)Joshi, Laurent and Bresson}]{Joshi2019}
\bibinfo{author}{Joshi, C.K.}, \bibinfo{author}{Laurent, T.},
  \bibinfo{author}{Bresson, X.}, \bibinfo{year}{2019}.
\newblock \bibinfo{title}{{On Learning Paradigms for the Travelling Salesman
  Problem}}.
\newblock \bibinfo{journal}{Advances in Neural Information Processing Systems}
  , \bibinfo{pages}{1--9}\URLprefix \url{http://arxiv.org/abs/1910.07210}.
\bibitem[{Karatzoglou et~al.(2018)Karatzoglou, Jablonski and
  Beigl}]{karatzoglou2018aseq2seq}
\bibinfo{author}{Karatzoglou, A.}, \bibinfo{author}{Jablonski, A.},
  \bibinfo{author}{Beigl, M.}, \bibinfo{year}{2018}.
\newblock \bibinfo{title}{A seq2seq learning approach for modeling semantic
  trajectories and predicting the next location}, in:
  \bibinfo{booktitle}{Proceedings of the 26th ACM SIGSPATIAL International
  Conference on Advances in Geographic Information Systems},
  \bibinfo{publisher}{Association for Computing Machinery},
  \bibinfo{address}{New York, NY, USA}. p. \bibinfo{pages}{528–531}.
\newblock \URLprefix \url{https://doi.org/10.1145/3274895.3274983},
  \DOIprefix\doi{10.1145/3274895.3274983}.
\bibitem[{Kool et~al.(2019)Kool, Van~Hoof and Welling}]{Kool2019}
\bibinfo{author}{Kool, W.}, \bibinfo{author}{Van~Hoof, H.},
  \bibinfo{author}{Welling, M.}, \bibinfo{year}{2019}.
\newblock \bibinfo{title}{{Attention, learn to solve routing problems!}}
\newblock \bibinfo{journal}{7th International Conference on Learning
  Representations, ICLR 2019} , \bibinfo{pages}{1--25}.
\bibitem[{K\"{u}{\c{c}}\"{u}ko{\u{g}}lu
  et~al.(2019)K\"{u}{\c{c}}\"{u}ko{\u{g}}lu, Dewil and Cattrysse}]{Kkolu2019}
\bibinfo{author}{K\"{u}{\c{c}}\"{u}ko{\u{g}}lu, {\.{I}}.},
  \bibinfo{author}{Dewil, R.}, \bibinfo{author}{Cattrysse, D.},
  \bibinfo{year}{2019}.
\newblock \bibinfo{title}{Hybrid simulated annealing and tabu search method for
  the electric travelling salesman problem with time windows and mixed charging
  rates}.
\newblock \bibinfo{journal}{Expert Systems with Applications}
  \bibinfo{volume}{134}, \bibinfo{pages}{279--303}.
\newblock \URLprefix \url{https://doi.org/10.1016/j.eswa.2019.05.037},
  \DOIprefix\doi{10.1016/j.eswa.2019.05.037}.
\bibitem[{Liang et~al.(2020)Liang, Du and Li}]{liang2020abstractive}
\bibinfo{author}{Liang, Z.}, \bibinfo{author}{Du, J.}, \bibinfo{author}{Li,
  C.}, \bibinfo{year}{2020}.
\newblock \bibinfo{title}{Abstractive social media text summarization using
  selective reinforced seq2seq attention model}.
\newblock \bibinfo{journal}{Neurocomputing} \bibinfo{volume}{410},
  \bibinfo{pages}{432--440}.
\bibitem[{Lim and Winkenbach(2019)}]{lim2019configuring}
\bibinfo{author}{Lim, S.F.W.}, \bibinfo{author}{Winkenbach, M.},
  \bibinfo{year}{2019}.
\newblock \bibinfo{title}{Configuring the last-mile in business-to-consumer
  e-retailing}.
\newblock \bibinfo{journal}{California Management Review} \bibinfo{volume}{61},
  \bibinfo{pages}{132--154}.
\bibitem[{Liu et~al.(2020)Liu, Jiang, Chen, Ye, He and Sun}]{LIU2020102070}
\bibinfo{author}{Liu, S.}, \bibinfo{author}{Jiang, H.}, \bibinfo{author}{Chen,
  S.}, \bibinfo{author}{Ye, J.}, \bibinfo{author}{He, R.},
  \bibinfo{author}{Sun, Z.}, \bibinfo{year}{2020}.
\newblock \bibinfo{title}{Integrating dijkstra’s algorithm into deep inverse
  reinforcement learning for food delivery route planning}.
\newblock \bibinfo{journal}{Transportation Research Part E: Logistics and
  Transportation Review} \bibinfo{volume}{142}, \bibinfo{pages}{102070}.
\newblock \URLprefix
  \url{https://www.sciencedirect.com/science/article/pii/S1366554520307213},
  \DOIprefix\doi{https://doi.org/10.1016/j.tre.2020.102070}.
\bibitem[{Liu et~al.(2018)Liu, Wang, Sha, Chang and Sui}]{liu2018table}
\bibinfo{author}{Liu, T.}, \bibinfo{author}{Wang, K.}, \bibinfo{author}{Sha,
  L.}, \bibinfo{author}{Chang, B.}, \bibinfo{author}{Sui, Z.},
  \bibinfo{year}{2018}.
\newblock \bibinfo{title}{Table-to-text generation by structure-aware seq2seq
  learning}, in: \bibinfo{booktitle}{Thirty-Second AAAI Conference on
  Artificial Intelligence}.
\bibitem[{Lu et~al.(2021)Lu, Rai, Chang, Knyazev, Yu, Shekhar, Taylor and
  Volkovs}]{lu2021context}
\bibinfo{author}{Lu, Y.}, \bibinfo{author}{Rai, H.}, \bibinfo{author}{Chang,
  J.}, \bibinfo{author}{Knyazev, B.}, \bibinfo{author}{Yu, G.},
  \bibinfo{author}{Shekhar, S.}, \bibinfo{author}{Taylor, G.W.},
  \bibinfo{author}{Volkovs, M.}, \bibinfo{year}{2021}.
\newblock \bibinfo{title}{Context-aware scene graph generation with seq2seq
  transformers}, in: \bibinfo{booktitle}{Proceedings of the IEEE/CVF
  International Conference on Computer Vision (ICCV)}, pp.
  \bibinfo{pages}{15931--15941}.
\bibitem[{Luong et~al.(2015)Luong, Pham and Manning}]{luong2015effective}
\bibinfo{author}{Luong, M.T.}, \bibinfo{author}{Pham, H.},
  \bibinfo{author}{Manning, C.D.}, \bibinfo{year}{2015}.
\newblock \bibinfo{title}{Effective approaches to attention-based neural
  machine translation}.
\newblock \bibinfo{journal}{arXiv preprint arXiv:1508.04025} .
\bibitem[{Ma et~al.(2019)Ma, Ge, He, Thaker and Drori}]{Ma2019}
\bibinfo{author}{Ma, Q.}, \bibinfo{author}{Ge, S.}, \bibinfo{author}{He, D.},
  \bibinfo{author}{Thaker, D.}, \bibinfo{author}{Drori, I.},
  \bibinfo{year}{2019}.
\newblock \bibinfo{title}{{Combinatorial Optimization by Graph Pointer Networks
  and Hierarchical Reinforcement Learning}} \URLprefix
  \url{http://arxiv.org/abs/1911.04936}.
\bibitem[{Matai et~al.(2010)Matai, Singh and Lal}]{Matai2010}
\bibinfo{author}{Matai, R.}, \bibinfo{author}{Singh, S.}, \bibinfo{author}{Lal,
  M.}, \bibinfo{year}{2010}.
\newblock \bibinfo{title}{Traveling salesman problem: an overview of
  applications, formulations, and solution approaches}, in:
  \bibinfo{booktitle}{Traveling Salesman Problem, Theory and Applications}.
  \bibinfo{publisher}{{InTech}}.
\newblock \URLprefix \url{https://doi.org/10.5772/12909},
  \DOIprefix\doi{10.5772/12909}.
\bibitem[{{McKinsey {\&} Company}(2021)}]{McKinseyCompany2021HowCountries}
\bibinfo{author}{{McKinsey {\&} Company}}, \bibinfo{year}{2021}.
\newblock \bibinfo{title}{{How e-commerce share of retail soared across the
  globe: A look at eight countries}}.
\newblock \URLprefix
  \url{https://www.mckinsey.com/featured-insights/coronavirus-leading-through-the-crisis/charting-the-path-to-the-next-normal/how-e-commerce-share-of-retail-soared-across-the-globe-a-look-at-eight-countries}.
\bibitem[{Merch{\'a}n et~al.(2022)Merch{\'a}n, Arora, Pachon, Konduri,
  Winkenbach, Parks and Noszek}]{merchan20222021}
\bibinfo{author}{Merch{\'a}n, D.}, \bibinfo{author}{Arora, J.},
  \bibinfo{author}{Pachon, J.}, \bibinfo{author}{Konduri, K.},
  \bibinfo{author}{Winkenbach, M.}, \bibinfo{author}{Parks, S.},
  \bibinfo{author}{Noszek, J.}, \bibinfo{year}{2022}.
\newblock \bibinfo{title}{2021 amazon last mile routing research challenge:
  Data set}.
\newblock \bibinfo{journal}{Transportation Science} .
\bibitem[{Mladenovi{\'{c}} et~al.(2012)Mladenovi{\'{c}}, Todosijevi{\'{c}} and
  Uro{\v{s}}evi{\'{c}}}]{Mladenovi2012}
\bibinfo{author}{Mladenovi{\'{c}}, N.}, \bibinfo{author}{Todosijevi{\'{c}},
  R.}, \bibinfo{author}{Uro{\v{s}}evi{\'{c}}, D.}, \bibinfo{year}{2012}.
\newblock \bibinfo{title}{An efficient {GVNS} for solving traveling salesman
  problem with time windows}.
\newblock \bibinfo{journal}{Electronic Notes in Discrete Mathematics}
  \bibinfo{volume}{39}, \bibinfo{pages}{83--90}.
\newblock \URLprefix \url{https://doi.org/10.1016/j.endm.2012.10.012},
  \DOIprefix\doi{10.1016/j.endm.2012.10.012}.
\bibitem[{Mo et~al.(2021)Mo, Zhao, Koutsopoulos and Zhao}]{mo2021individual}
\bibinfo{author}{Mo, B.}, \bibinfo{author}{Zhao, Z.},
  \bibinfo{author}{Koutsopoulos, H.N.}, \bibinfo{author}{Zhao, J.},
  \bibinfo{year}{2021}.
\newblock \bibinfo{title}{Individual mobility prediction in mass transit
  systems using smart card data: An interpretable activity-based hidden markov
  approach}.
\newblock \bibinfo{journal}{IEEE Transactions on Intelligent Transportation
  Systems} \bibinfo{volume}{23}, \bibinfo{pages}{12014--12026}.
\bibitem[{{Pitney Bowes}(2020)}]{PitneyBowes2020PitneyIndex}
\bibinfo{author}{{Pitney Bowes}}, \bibinfo{year}{2020}.
\newblock \bibinfo{title}{{Pitney Bowes Parcel Shipping Index}}.
\newblock \URLprefix \url{https://www.pitneybowes.com/us/shipping-index.html}.
\bibitem[{{postnord}(2021)}]{postnord2021E-commerceEurope}
\bibinfo{author}{{postnord}}, \bibinfo{year}{2021}.
\newblock \bibinfo{title}{{E-commerce in Europe 2020 - How the pandemic is
  changing e-commerce in Europe}}.
\newblock \bibinfo{type}{Technical Report}.
\newblock \URLprefix
  \url{https://www.postnord.se/siteassets/pdf/rapporter/e-commerce-in-europe-2020.pdf}.
\bibitem[{Purkayastha et~al.(2020)Purkayastha, Chakraborty, Saha and
  Mukhopadhyay}]{Purkayastha2020}
\bibinfo{author}{Purkayastha, R.}, \bibinfo{author}{Chakraborty, T.},
  \bibinfo{author}{Saha, A.}, \bibinfo{author}{Mukhopadhyay, D.},
  \bibinfo{year}{2020}.
\newblock \bibinfo{title}{Study and analysis of various heuristic algorithms
  for solving travelling salesman problem{\textemdash}a survey}, in:
  \bibinfo{booktitle}{Advances in Intelligent Systems and Computing}.
  \bibinfo{publisher}{Springer Singapore}, pp. \bibinfo{pages}{61--70}.
\newblock \URLprefix \url{https://doi.org/10.1007/978-981-15-2188-1_5},
  \DOIprefix\doi{10.1007/978-981-15-2188-1_5}.
\bibitem[{Pytorch(2021)}]{lstm_pytorch}
\bibinfo{author}{Pytorch}, \bibinfo{year}{2021}.
\newblock \bibinfo{title}{Pytorch {LSTM} document}.
\newblock \URLprefix
  \url{https://pytorch.org/docs/stable/generated/torch.nn.LSTM.html}.
\bibitem[{Ren et~al.(2020)Ren, Choi, Lee and Lin}]{REN2020101834}
\bibinfo{author}{Ren, S.}, \bibinfo{author}{Choi, T.M.}, \bibinfo{author}{Lee,
  K.M.}, \bibinfo{author}{Lin, L.}, \bibinfo{year}{2020}.
\newblock \bibinfo{title}{Intelligent service capacity allocation for
  cross-border-e-commerce related third-party-forwarding logistics operations:
  A deep learning approach}.
\newblock \bibinfo{journal}{Transportation Research Part E: Logistics and
  Transportation Review} \bibinfo{volume}{134}, \bibinfo{pages}{101834}.
\newblock \URLprefix
  \url{https://www.sciencedirect.com/science/article/pii/S1366554519311688},
  \DOIprefix\doi{https://doi.org/10.1016/j.tre.2019.101834}.
\bibitem[{Rose et~al.(2016)Rose, Mollenkopf, Autry and
  Bell}]{Rose2016ExploringProviders}
\bibinfo{author}{Rose, W.J.}, \bibinfo{author}{Mollenkopf, D.A.},
  \bibinfo{author}{Autry, C.}, \bibinfo{author}{Bell, J.},
  \bibinfo{year}{2016}.
\newblock \bibinfo{title}{{Exploring urban institutional pressures on logistics
  service providers}}.
\newblock \bibinfo{journal}{International Journal of Physical Distribution {\&}
  Logistics Management} \bibinfo{volume}{46}.
\newblock \DOIprefix\doi{10.1108/09600035199500001}.
\bibitem[{Salman et~al.(2020)Salman, Ekstedt and Damaschke}]{salman2020branch}
\bibinfo{author}{Salman, R.}, \bibinfo{author}{Ekstedt, F.},
  \bibinfo{author}{Damaschke, P.}, \bibinfo{year}{2020}.
\newblock \bibinfo{title}{Branch-and-bound for the precedence constrained
  generalized traveling salesman problem}.
\newblock \bibinfo{journal}{Operations Research Letters} \bibinfo{volume}{48},
  \bibinfo{pages}{163--166}.
\bibitem[{da~Silva and Urrutia(2010)}]{daSilva2010}
\bibinfo{author}{da~Silva, R.F.}, \bibinfo{author}{Urrutia, S.},
  \bibinfo{year}{2010}.
\newblock \bibinfo{title}{A general {VNS} heuristic for the traveling salesman
  problem with time windows}.
\newblock \bibinfo{journal}{Discrete Optimization} \bibinfo{volume}{7},
  \bibinfo{pages}{203--211}.
\newblock \URLprefix \url{https://doi.org/10.1016/j.disopt.2010.04.002},
  \DOIprefix\doi{10.1016/j.disopt.2010.04.002}.
\bibitem[{Snoeck and Winkenbach(2021)}]{snoeck2021discrete}
\bibinfo{author}{Snoeck, A.}, \bibinfo{author}{Winkenbach, M.},
  \bibinfo{year}{2021}.
\newblock \bibinfo{title}{A discrete simulation-based optimization algorithm
  for the design of highly responsive last-mile distribution networks}.
\newblock \bibinfo{journal}{Transportation Science} .
\bibitem[{Sutskever et~al.(2014)Sutskever, Vinyals and
  Le}]{Sutskever2014SequenceNetworks}
\bibinfo{author}{Sutskever, I.}, \bibinfo{author}{Vinyals, O.},
  \bibinfo{author}{Le, Q.V.}, \bibinfo{year}{2014}.
\newblock \bibinfo{title}{{Sequence to sequence learning with neural
  networks}}, in: \bibinfo{booktitle}{Advances in Neural Information Processing
  Systems}, pp. \bibinfo{pages}{3104--3112}.
\bibitem[{Traub et~al.(2021)Traub, Vygen and Zenklusen}]{Traub2021}
\bibinfo{author}{Traub, V.}, \bibinfo{author}{Vygen, J.},
  \bibinfo{author}{Zenklusen, R.}, \bibinfo{year}{2021}.
\newblock \bibinfo{title}{Reducing path {TSP} to {TSP}}.
\newblock \bibinfo{journal}{{SIAM} Journal on Computing} ,
  \bibinfo{pages}{STOC20--24--STOC20--53}\URLprefix
  \url{https://doi.org/10.1137/20m135594x}, \DOIprefix\doi{10.1137/20m135594x}.
\bibitem[{{United Nations Department of Economic and Social
  Affairs}(2019)}]{UnitedNationsDepartmentofEconomicandSocialAffairs2019WorldST/ESA/SER.A/420}
\bibinfo{author}{{United Nations Department of Economic and Social Affairs}},
  \bibinfo{year}{2019}.
\newblock \bibinfo{title}{{World Urbanization Prospects: The 2018 Revision
  (ST/ESA/SER.A/420)}}.
\newblock \bibinfo{type}{Technical Report}. \bibinfo{address}{New York: United
  Nations}.
\newblock \URLprefix
  \url{https://population.un.org/wup/Publications/Files/WUP2018-Report.pdf}.
\bibitem[{{US Census Bureau}(2021)}]{USCensusBureau2021Quarterly2021}
\bibinfo{author}{{US Census Bureau}}, \bibinfo{year}{2021}.
\newblock \bibinfo{title}{{Quarterly e-commerce retail sales 2nd quarter
  2021}}.
\newblock \bibinfo{type}{Technical Report}. U.S. Census Bureau of the
  Department of Commerce.
\newblock \URLprefix
  \url{http://www2.census.gov/retail/releases/historical/ecomm/07q4.pdf}.
\bibitem[{Vaswani et~al.(2017)Vaswani, Shazeer, Parmar, Uszkoreit, Jones,
  Gomez, Kaiser and Polosukhin}]{Vaswani2017}
\bibinfo{author}{Vaswani, A.}, \bibinfo{author}{Shazeer, N.},
  \bibinfo{author}{Parmar, N.}, \bibinfo{author}{Uszkoreit, J.},
  \bibinfo{author}{Jones, L.}, \bibinfo{author}{Gomez, A.N.},
  \bibinfo{author}{Kaiser, L.}, \bibinfo{author}{Polosukhin, I.},
  \bibinfo{year}{2017}.
\newblock \bibinfo{title}{{Attention Is All You Need}}.
\newblock \bibinfo{journal}{Advances in Neural Information Processing Systems}
  \bibinfo{volume}{2017-Decem}, \bibinfo{pages}{5999--6009}.
\newblock \URLprefix \url{http://arxiv.org/abs/1706.03762},
  \href{http://arxiv.org/abs/1706.03762}{\tt arXiv:1706.03762}.
\bibitem[{Vinyals et~al.(2016)Vinyals, Bengio and Kudlur}]{Vinyals2016}
\bibinfo{author}{Vinyals, O.}, \bibinfo{author}{Bengio, S.},
  \bibinfo{author}{Kudlur, M.}, \bibinfo{year}{2016}.
\newblock \bibinfo{title}{{Order matters: Sequence to sequence for sets}}.
\newblock \bibinfo{journal}{4th International Conference on Learning
  Representations, ICLR 2016 - Conference Track Proceedings} ,
  \bibinfo{pages}{1--11}.
\bibitem[{Vinyals et~al.(2015)Vinyals, Meire and Navdeep}]{Oriol2015}
\bibinfo{author}{Vinyals, O.}, \bibinfo{author}{Meire, F.},
  \bibinfo{author}{Navdeep, J.}, \bibinfo{year}{2015}.
\newblock \bibinfo{title}{{Pointer Networks}}.
\newblock \bibinfo{journal}{Advances in Neural Information Processing Systems}
  , \bibinfo{pages}{1--9}.
\bibitem[{Wang et~al.(2020a)Wang, Cao, Chen, Peng and
  Huang}]{wang2020seqst-gan}
\bibinfo{author}{Wang, S.}, \bibinfo{author}{Cao, J.}, \bibinfo{author}{Chen,
  H.}, \bibinfo{author}{Peng, H.}, \bibinfo{author}{Huang, Z.},
  \bibinfo{year}{2020}a.
\newblock \bibinfo{title}{Seqst-gan: Seq2seq generative adversarial nets for
  multi-step urban crowd flow prediction} \bibinfo{volume}{6}.
\newblock \URLprefix \url{https://doi.org/10.1145/3378889},
  \DOIprefix\doi{10.1145/3378889}.
\bibitem[{Wang et~al.(2020b)Wang, Mo and Zhao}]{Wang2020DeepFunctions}
\bibinfo{author}{Wang, S.}, \bibinfo{author}{Mo, B.}, \bibinfo{author}{Zhao,
  J.}, \bibinfo{year}{2020}b.
\newblock \bibinfo{title}{{Deep neural networks for choice analysis:
  Architecture design with alternative-specific utility functions}}.
\newblock \bibinfo{journal}{Transportation Research Part C: Emerging
  Technologies} \bibinfo{volume}{112}, \bibinfo{pages}{234--251}.
\newblock \DOIprefix\doi{10.1016/J.TRC.2020.01.012}.
\bibitem[{Winkenbach et~al.(2021)Winkenbach, Parks and
  Noszek}]{Winkenbach2021TechnicalChallenge}
\bibinfo{author}{Winkenbach, M.}, \bibinfo{author}{Parks, S.},
  \bibinfo{author}{Noszek, J.}, \bibinfo{year}{2021}.
\newblock \bibinfo{title}{{Technical Proceedings of the Amazon Last Mile
  Routing Research Challenge}} \URLprefix
  \url{https://dspace.mit.edu/handle/1721.1/131235}.
\bibitem[{Wu et~al.(2020)Wu, Zhuang, Xu, Zhang and Chen}]{wu2020pq-net}
\bibinfo{author}{Wu, R.}, \bibinfo{author}{Zhuang, Y.}, \bibinfo{author}{Xu,
  K.}, \bibinfo{author}{Zhang, H.}, \bibinfo{author}{Chen, B.},
  \bibinfo{year}{2020}.
\newblock \bibinfo{title}{Pq-net: A generative part seq2seq network for 3d
  shapes}, in: \bibinfo{booktitle}{Proceedings of the IEEE/CVF Conference on
  Computer Vision and Pattern Recognition (CVPR)}.
\bibitem[{Xu et~al.(2017)Xu, Wang, Zhu and Huang}]{xu2017seq2seq}
\bibinfo{author}{Xu, Z.}, \bibinfo{author}{Wang, S.}, \bibinfo{author}{Zhu,
  F.}, \bibinfo{author}{Huang, J.}, \bibinfo{year}{2017}.
\newblock \bibinfo{title}{Seq2seq fingerprint: An unsupervised deep molecular
  embedding for drug discovery}, in: \bibinfo{booktitle}{Proceedings of the 8th
  ACM International Conference on Bioinformatics, Computational Biology,and
  Health Informatics}, \bibinfo{publisher}{Association for Computing
  Machinery}, \bibinfo{address}{New York, NY, USA}. p.
  \bibinfo{pages}{285–294}.
\newblock \URLprefix \url{https://doi.org/10.1145/3107411.3107424},
  \DOIprefix\doi{10.1145/3107411.3107424}.
\bibitem[{Yuan et~al.(2020)Yuan, Cattaruzza, Ogier and Semet}]{yuan2020branch}
\bibinfo{author}{Yuan, Y.}, \bibinfo{author}{Cattaruzza, D.},
  \bibinfo{author}{Ogier, M.}, \bibinfo{author}{Semet, F.},
  \bibinfo{year}{2020}.
\newblock \bibinfo{title}{A branch-and-cut algorithm for the generalized
  traveling salesman problem with time windows}.
\newblock \bibinfo{journal}{European Journal of Operational Research}
  \bibinfo{volume}{286}, \bibinfo{pages}{849--866}.
\bibitem[{Zax(2013)}]{fastcompany2013UPS}
\bibinfo{author}{Zax, D.}, \bibinfo{year}{2013}.
\newblock \bibinfo{title}{Brown down: {UPS} drivers vs. the {UPS} algorithm}.
\newblock \URLprefix
  \url{https://www.fastcompany.com/3004319/brown-down-ups-drivers-vs-ups-algorithm}.
  \bibinfo{note}{accessed February 14, 2023}.
\bibitem[{Zhang et~al.(2019)Zhang, Li, Wang, Fang and Xiao}]{zhang2019abstract}
\bibinfo{author}{Zhang, Y.}, \bibinfo{author}{Li, D.}, \bibinfo{author}{Wang,
  Y.}, \bibinfo{author}{Fang, Y.}, \bibinfo{author}{Xiao, W.},
  \bibinfo{year}{2019}.
\newblock \bibinfo{title}{Abstract text summarization with a convolutional
  seq2seq model}.
\newblock \bibinfo{journal}{Applied Sciences} \bibinfo{volume}{9}.
\newblock \URLprefix \url{https://www.mdpi.com/2076-3417/9/8/1665},
  \DOIprefix\doi{10.3390/app9081665}.
\bibitem[{Zhang et~al.(2020)Zhang, Li and Zhang}]{zhang2020short-term}
\bibinfo{author}{Zhang, Y.}, \bibinfo{author}{Li, Y.}, \bibinfo{author}{Zhang,
  G.}, \bibinfo{year}{2020}.
\newblock \bibinfo{title}{Short-term wind power forecasting approach based on
  seq2seq model using nwp data}.
\newblock \bibinfo{journal}{Energy} \bibinfo{volume}{213},
  \bibinfo{pages}{118371}.
\newblock \URLprefix
  \url{https://www.sciencedirect.com/science/article/pii/S036054422031478X},
  \DOIprefix\doi{https://doi.org/10.1016/j.energy.2020.118371}.

\end{thebibliography}

\end{document}